\pdfoutput=1

\documentclass[11pt]{article}

\usepackage[preprint]{acl}

\usepackage{times}
\usepackage{latexsym}

\usepackage[T1]{fontenc}

\usepackage[utf8]{inputenc}

\usepackage{xspace}
\usepackage{microtype}
\usepackage{hyperref}
\usepackage{url}
\usepackage{amsmath} 
\usepackage{enumitem}
\usepackage{booktabs}
\usepackage{algorithm}
\usepackage{algpseudocode}
\usepackage{listings}
\usepackage{xcolor}
\usepackage{multirow}
\usepackage{graphicx}
\usepackage{booktabs}
\usepackage{color}
\usepackage{colortbl}
\usepackage{wrapfig}
\usepackage{tikz}
\usetikzlibrary{tikzmark}
\usepackage{longtable}
\usepackage{array}
\usepackage{setspace}
\usepackage{fancybox}

\definecolor{myred}{RGB}{255, 99, 71}
\definecolor{mygreen}{RGB}{0, 255, 0}
\definecolor{darkblue}{rgb}{0, 0, 0.5}
\usepackage{svg}
\hypersetup{colorlinks=true, citecolor=darkblue, linkcolor=darkblue, urlcolor=darkblue}

\usepackage{inconsolata}

\newcommand{\mname}{{\sc Fanno}\xspace}

\title{\mname: Augmenting High-Quality Instruction Data with \\ Open-Sourced LLMs Only}

\author{
  He Zhu$^{1,2}$,  Junyou Su$^{1}$, Tianle Lun$^1$, Yicheng Tao$^1$ \\ Wenjia Zhang$^2$, Zipei Fan$^3$, Guanhua Chen$^1\thanks{Corresponding author}$ \\
  $^1$ Southern University of Science and Technology,
  $^2$ Peking University,
  $^3$ Jilin University 
}

\begin{document}
\maketitle
\begin{abstract}
Instruction fine-tuning stands as a crucial advancement in leveraging large language models (LLMs) for enhanced task performance. However, the annotation of instruction datasets has traditionally been expensive and laborious, often relying on manual annotations or costly API calls of proprietary LLMs. To address these challenges, we introduce \mname, a fully autonomous, open-sourced framework that revolutionizes the annotation process without the need for pre-existing annotated data. Utilizing a Mistral-7b-instruct model, \mname efficiently produces diverse and high-quality datasets through a structured process involving document pre-screening, instruction generation, and response generation. Experiments on Open LLM Leaderboard and AlpacaEval benchmark show that the \mname can generate high-quality data with diversity and complexity for free, comparable to human-annotated or cleaned datasets like Alpaca-GPT4-Cleaned.
\end{abstract}

\section{Introduction}
Large language models (LLMs) have made significant contributions across numerous fields~\citep{zheng2024step,wang2024llms,wettig2024qurating,fan2023large}. Instruction tuning \citep{ouyang2022training} enhances the model's general capabilities for novel tasks and improves their adherence to directives. However, the development of human-annotated instruction data is prohibitively expensive, and often results in suboptimal outcomes~\citep{srivastava2022beyond,dolly2023}. This is primarily due to the annotators' cognitive limitations, which hinder achieving a balanced dataset in terms of diversity, complexity, and quality~\citep{srivastava2022beyond,dolly2023}. 
Previous works explore the automatic LLM-based annotation of instruction data, with advanced proprietary models \citep{wang2022self,xu2023wizardlm} or models trained with seed response-query pairs \citep{li2024selfalignment,lou2024muffin}. 
Nevertheless, these approaches often depend on costly APIs (ChatGPT/GPT-4) or require manually crafted seed datasets.
Recent studies \citep{zheng2024kun,yehudai2024genie,press2023measuring} aim to construct instruction datasets from scratch; however, the strategies to balance the diversity, complexity, and quality \citep{liu2023what} of annotated instruction data are less explored.

Addressing these challenges, we introduce \mname (Free ANNOtator), a freely accessible framework specifically designed for automatic high-quality instruction annotation. This framework methodically breaks down the annotation process into three distinct phases: document pre-screen, instruction generation, and response generation.  It utilizes curated tagging, UCB(Upper Confidence Bound) bootstrapping iterations, and filtering techniques to enhance the diversity and complexity of the generated instructions. Empirical evidence on Open LLM Leaderboard and AlpacaEval benchmark confirm the framework's efficacy on two 7B LLMs. The resulting dataset is virtually indistinguishable from those refined datasets like \textbf{Alpaca-GPT4-Cleaned}, marking a significant stride in instruction data development\footnote{Our code, data, and model will be made public.}.




\section{Related Work}

\paragraph{Instruction Data Generation} Two main approaches have been explored for instruction data creation: (1) \textbf{Human Annotation}, which leverages human expertise to design prompts and collect multi-task datasets spanning various categories~\citep{srivastava2022beyond,dolly2023}. While producing high-quality data, manual annotation is effort-intensive and costly, especially for devising complex textual instructions. (2) \textbf{LLM Synthetic Data Generation} 
Recent research increasingly favors harnessing the creative capabilities of LLMs, such as GPT-4~\citep{openai2023gpt4}, over human input for creating instruction-following datasets~\citep{koala_blogpost_2023,vicuna2023}. \textsc{Alpaca}~\citep{alpaca} and \textsc{AlpacaGPT}~\citep{peng2023gpt4llm} have also utilized more powerful LLMs to enhance data quality. Another line of research involves generating task instructions from ``seeds'' and filtering~\citep{wu2023lamini}. For example, \textsc{WizardLM}~\citep{xu2023wizardlm} employed an instruction evolution paradigm to increase seed instruction complexity, while \textsc{self-instruct}~\citep{wang2022self} used human-annotated instructions as demonstrations to guide LLMs in the instruction evolution process. Humpback~\citep{li2024selfalignment} generates instructions using vast amounts of unlabeled web text. These datasets are costly, either in terms of labor or proprietary model expenses. In contrast, \mname maintains high instructional quality autonomously, utilizing open-source models efficiently with just a 7B model size.

\paragraph{Instruction Tuning}
Instruction tuning involves training LLMs on extensive upstream task datasets with instructions, followed by enabling the generalized ability to new, unseen downstream tasks via new instructions~\citep{ouyang2022training,chung2022scaling}. This technique is widely acknowledged as essential for activating LLMs to adhere to human conversational norms~\citep{mishra2022cross}.
Instruction tuning has empowered various domain-specific or task-specific LLMs~\citep{jiang2023tigerscore, xu2023wizardlm}, and curating diverse, high-quality upstream instruction dataset has become a pivotal step for successful instruction tuning~\citep{wang2023far,lou2023prompt}.
Moreover, instruction tuning also bolsters cross-task general capabilities~\citep{sanh2021multitask,wang-etal-2022-super}, encompassing a more comprehensive array of general tasks, notably incorporating input from users of language models~\citep{ouyang2022training,peng2023gpt4llm}.

\paragraph{Data Quality Enhancement}
Related works in the field of enhancing data quality have focused on several key aspects such as instruction difficulty, diversity, and correctness. \textsc{Humpback}~\citep{li2024selfalignment} and \textsc{Kun}~\citep{zheng2024kun} utilize language model's capability in combination with tailored prompts for data filtering. In Addition, initiatives like \textsc{Genie}~\citep{yehudai2024genie} and \textsc{MoDS}~\citep{du2023mods} utilize specialized open-source LLMs for data filtering tasks. \textsc{Deita}~\citep{liu2023what}, \textsc{PlanGPT}~\citep{zhu2024plangpt} and similar approaches utilize fine-tuned large models to score the data for quality assessment. Moreover, efforts like \textsc{Orca-Math}~\citep{mitra2024orcamath} and \textsc{Reflection-tuning}~\citep{Li2023ReflectionTuningDR} employ collaborative approaches with multiple LMs and self-reflection to enhance data quality.

\section{\mname Framework} \label{framework}
\begin{figure*}[h]
    \centering
    \includegraphics[width=0.9\textwidth]{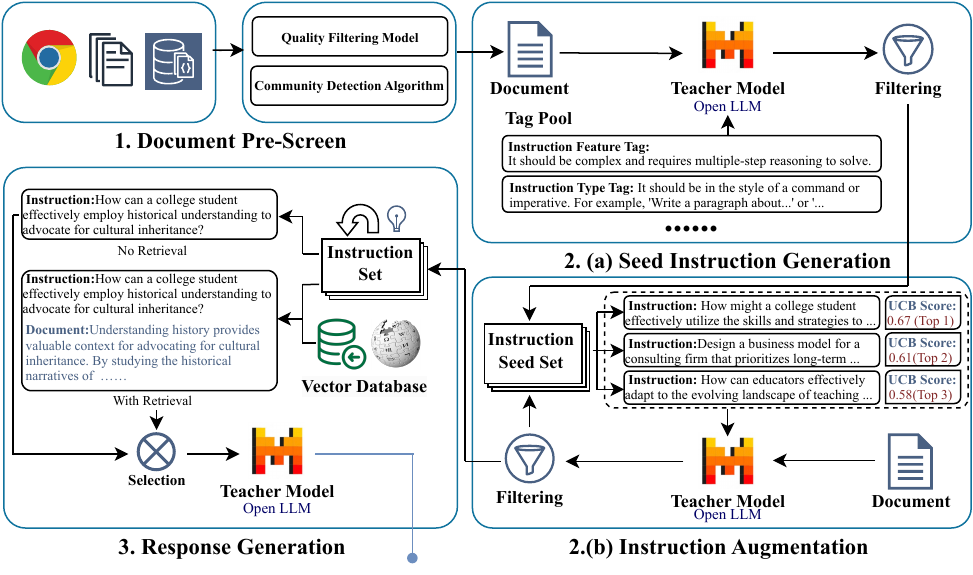}
    \caption{Overview of \mname framework. ~\textbf{(1) Document Pre-Screen}: We process the unlabeled text data with filters and community detection algorithm. ~\textbf{(2a) Seed Instruction Generation}: \mname generates seed instructions from pre-screened documents with diverse task types and difficulty levels through a tag pool. ~\textbf{(2b) Instruction Augmentation}: New instructions are augmented conditioned on the documents and few-shot examples selected from the seed instructions with the UCB algorithm. ~\textbf{(3) Response Generation}: The responses to instructions are generated directly by the teacher LLM or based on the concatenation of the corresponding document and retrieved document. }
    \label{fig:fanno}
\end{figure*}

The \mname framework aims to annotate diverse, complex, and faithful instruction data with only free open-sourced LLMs. As depicted in Figure~\ref{fig:fanno}, \mname consists of three pivotal steps: document pre-screen, instruction generation, and response generation. 

\subsection{Document Pre-Screen}

The \mname framework annotates instruction data from web corpus, textbooks, etc. The document pre-screening stage initially includes segmentation, deduplication, and length-based filtering. Further filtering employs a teacher LLM and a fast community detection algorithm to enhance correctness and diversity. 

The LLM-based filter addresses ambiguous content, privacy concerns, and advertisements (see Appendix~\ref{appendix:text_filtering}). To reduce data volume while maintaining diversity, we cluster instruction embeddings using a fast community detection algorithm, similar to SentenceTransformer \citep{reimers-2019-sentence-bert}, based on a predefined similarity threshold. This approach prioritizes larger, non-overlapping communities (details are in Appendix~\ref{appendix:fast_community_detection_algorithm} and  Algorithm~\ref{algorithm:fast_community_detection}). 

The pre-screen phase balances processing speed and precision, prioritizing efficiency. In our experiments, the pre-screen stage filters and keeps 6\% of the original raw data.

\subsection{Instruction Generation}
At this stage, \mname adopts a bootstrapping approach to generate instructions from pre-screened documents, streamlining the process into two distinct phases: seed instruction generation and instruction augmentation.

\paragraph{Step 1: Seed Instruction Generation} This step produces a set of diverse instructions as the initial seeds. Diversity is promoted from two perspectives: \textbf{Task Types} and \textbf{Difficulty Levels}, for which we have manually created corresponding tags (see Appendix~\ref{sec:app-seed-gen}). For each document, we traverse all combinations of task types and instruction difficulty levels to generate seed instructions. An LLM-based filter (see Table~\ref{table:text_quality_prompt_2} in the appendix) is then employed to ensure the quality of the seed instruction data. We sample 200 documents for instruction generation and obtain around 1k instructions as the seed pool \( S \).

\paragraph{Step 2: Instruction Augmentation} The diversity of the instructions in \( S \) is inherently limited. To promote the diversity of newly generated instructions, we designed a prompt template called \textbf{Think Different} (see Appendix~\ref{table:ucb_think_diff}), which diverges from the traditional example-followed template used in self-instruct (see Appendix~\ref{sec:app-self-instruct-gen}). This template encourages the teacher model to generate high-quality instructions that emulate the quality of the examples but differ in format (task types, questioning styles, etc.). Additionally, a document is inputted into this template to ensure the generated instructions are consistent with or extended from this document.


The quality of the examples is, therefore, paramount. Instead of randomly selecting examples, we prioritize extracting higher-quality ones, assuming that instruction length correlates with quality. To avoid suboptimal convergence, the \textbf{UCB} (Upper Confidence Bound)~\citep{robbins1951stochastic} score is used to enhance the exploration of new instructions. Each seed data is scored as \( UCB(s) = \bar{x}_s + C\sqrt{\frac{2 \ln N}{n_s}} \). Here, \(\bar{x}_s\) is the seed's average quality, \(N\) is the total iterations, and \(C\) is a constant. The score promotes high-quality and less frequently selected seeds, with \(C\) balancing these objectives. In each iteration, we select \( k \) seeds with the highest UCB scores, effectively trade-off between exploration and exploitation. We compare UCB and random sampling in an ablation study. The detailed algorithm can be found in Appendix~\ref{appendix:UCB}.

\subsection{Response Generation} \label{sec:response_generation}
At this stage, the response to each instruction is generated by prompting the teacher LLM either with empty context or a retrieved document. We propose to apply retrieval augmented generation (RAG) and incorporate the corresponding document to provide additional information for response generation. These documents are concatenated to serve as the relevant context.
For all generations, the teacher LLM is prompted to generate responses under the above two different conditions. Then, we use the LLM itself to select the response with better quality. The prompt templates in Appendix~\ref{appendix:response_generate} and Table~\ref{table:faithfulness_evaluation_prompt} are used for response generation and selection, respectively. 


\subsection{Discussion}
To produce diverse and complex high-quality instruction data, \mname utilizes tags for difficulty balancing, iteratively selects high-quality data via UCB bootstrap, and ensures diversity through iterative instruction filtering. We generated strongly generalized data independent of the original text through carefully crafted prompts. To ensure fidelity between instructions and responses, information is supplemented using RAG and a Teacher model. Detailed discussions are in Section~\ref{analyses}.

\section{Experiment}
\subsection{Experiment setup}
\paragraph{Unlabeled Text Data}  We use the \textsc{Falcon Refined Web} corpus\footnote{\url{https://huggingface.co/datasets/tiiuae/falcon-refinedweb}}~\citep{penedo2023refinedweb}, a large web-based corpus dataset including 600 billion tokens, as our unlabeled data. We directly selected the first 500k documents for input to the Document Pre-Screen stage.


\paragraph{Models and Training Details}
We choose the Mistral-7b-instruct-v0.2~\citep{jiang2023mistral} for data annotation in all experiments. We perform supervised instruction tuning using LoRA~\citep{hu2021lora} with the pretrained LLaMA-2-7b-base model \citep{touvron2023llama} and Mistral-7b-base model~\citep{jiang2023mistral}. The model after instruction tuning with the data annotated with \mname framework is referred to as \mname. Detailed configuration can be viewed in Appendix~\ref{appendix:setting}.


\paragraph{Baselines} We compare \mname with models finetuned with other instruction datasets. The baseline details are in Appendix~\ref{appendix:baseline}.
The datasets include Alpaca-52k~\citep{alpaca}, Alpaca-GPT4~\citep{peng2023gpt4llm}, Alpaca-Cleaned, LIMA~\citep{zhou2023lima}, WizardLM-70k~\citep{xu2023wizardlm}, and Muffin~\citep{lou2024muffin}. Alpaca-52k and Alpaca-GPT4, each with 52,002 samples, use Text-Davinci-003 and GPT-4 for annotations. Alpaca-Cleaned refines Alpaca-GPT4 to 51,760 samples filtered instructions with hallucination errors or invalid outputs. LIMA offers 1,000 manually selected diverse prompts and responses. WizardLM-70k and Muffin, both using ChatGPT or GPT-4 annotations, focus on 70,000 and 68,000 high-quality samples, respectively. The self-augmented dataset of Humpback is also comparatively ensured to be fair.

\subsection{Evaluation}
\paragraph{Open LLM Leaderboard} The Huggingface Open LLM Leaderboard\footnote{\url{https://huggingface.co/spaces/HuggingFaceH4/open_llm_leaderboard}}~\citep{open-llm-leaderboard} stands as a unified framework designed to evaluate generative language models across a wide array of diverse evaluation tasks. It encompasses key benchmarks such as ARC~\citep{clark2018think}, HellaSwag~\citep{zellers2019hellaswag}, MMLU~\citep{hendrycks2021measuring}, and TruthfulQA~\citep{lin2022truthfulqa}. We utilize the lm-evaluation-harness toolkit\footnote{\url{https://github.com/EleutherAI/lm-evaluation-harness}} \citep{eval-harness} for evaluating different models to maintain consistency with the official setup. 

\paragraph{AlpacaEval 2.0} AlpacaEval Benchmark~\citep{alpaca_eval} is an automated evaluation framework based on a annotation model(GPT-4). By comparing responses generated by two different models for the same set of 805 prompts, AlpacaEval computes the pairwise win rate, automating the evaluation process. 

\paragraph{Human Evaluation}\label{sec:evaluation} 
We employed manual annotation by multiple experts to identify the complexity of instructions, specifically categorized into three tiers: (0 Unanswerable, 1 Easy, 2 Expert). The detailed information for each tier is provided in Appendix~\ref{appendix:human_eval_complexity}. The evaluation results are presented in Table~\ref{tab:human_test_results} and discussed in Section~\ref{sec:complex_discuss}.

\paragraph{MT-Bench} The MT-Bench (Multi-turn Benchmark)~\citep{zheng2024judging} is aimed at assessing the conversational and instruction-following abilities of LLMs. It comprises 80 multi-turn questions, and GPT-4 is utilized as an automated evaluator, scoring chatbot responses on a scale of 1 to 10, with methods in place to minimize bias and enhance the reliability of the assessments.


\begin{table*}[!t]
  \centering
    \resizebox{2 \columnwidth}{!}{
  \begin{tabular}{lcrrrrr}
    \toprule
    \textbf{Model} & \textbf{Data Size} & \textbf{ARC} & \textbf{HellaSwag} & \textbf{MMLU} & \textbf{TruthfulQA} & \textbf{Average}  \\
    \midrule
    \multicolumn{7}{c}{\textbf{Open-sourced Models based on LLaMA-2}}\\
    LLaMA-2-Base & -- & 54.10 & 78.71 & 45.80 & 38.96 & 50.76  \\
    LLaMA-2-Chat & -- & 54.10 & 78.65 & 45.69 & 44.59 & 55.76 \\
    LLaMA-2 + Alpaca-52k & 52k & 54.78 & 78.17 & 46.65 & 41.43 & 55.26 \\
    LLaMA-2 + Alpaca-GPT4 & 52k  &  \cellcolor{myred!25}56.66 & 78.78 & 46.96 &  \cellcolor{myred!25}51.02 & \cellcolor{mygreen!25}58.35 \\ 
    LLaMA-2 + Alpaca-Cleaned & 51.8k  & \cellcolor{mygreen!25}56.40 &  \cellcolor{myred!25}80.16 & \cellcolor{mygreen!25}47.02 & 50.53 &  \cellcolor{myred!25}58.53 \\ 
    LLaMA-2 + LIMA & 1k & 54.61 & 79.21 & 45.79 & 41.32 & 55.23 \\
    LLaMA-2 + WizardLM-70k & 65k & 54.01 & 78.66 & 45.61 & 38.99 & 54.32 \\  
    LLaMA-2 + Muffin & 68k & 54.10 & 76.97 &  \cellcolor{myred!25}47.12 & 43.51 & 55.42 \\
    LLaMA-2 + \mname  & 16k  & 55.63  & \cellcolor{mygreen!25}79.45 & 46.84 & \cellcolor{mygreen!25}51.01 & 58.23  \\ 
    \midrule
    \multicolumn{7}{c}{\textbf{Open-sourced Models based on Mistral-7B}}\\
    Mistral-7B-Instruct-v0.2 & -- & 59.39  & \cellcolor{mygreen!25}84.33   & 59.28  & \cellcolor{myred!25}66.79 & \cellcolor{myred!25}67.45 \\
    Mistral-7B-Base-v0.1 & -- & 60.84 &  83.31 & \cellcolor{myred!25}62.42  & 42.59 & 62.29 \\
    Mistral-7B-Base + Alpaca-GPT4 & 52k & 63.65 & 82.18 & 59.29 & 43.98 & 62.29 \\
    Mistral-7B-Base + Alpaca-Cleaned & 51.8K &  \cellcolor{myred!25} 64.51 & 83.68 & 59.76 & 52.00 & 64.99 \\
    Mistral-7B-Base + \mname & 16k & \cellcolor{mygreen!25}64.16 & \cellcolor{myred!25}85.08 & \cellcolor{mygreen!25}60.79 & \cellcolor{mygreen!25}52.16 & \cellcolor{mygreen!25}65.55  \\
    \bottomrule
    \end{tabular}
   }
  \caption{Open LLM Leaderboard results evaluated with the lm-evalution-harness toolkit.
  Data size represents the number of samples in the instruction data.}
  \label{tab:openllm1}
  \vspace{-10pt}
\end{table*}

\begin{table}[h]
\centering

\scalebox{0.7}{
\begin{tabular}{lcc}
\toprule
\textbf{Model} & \textbf{MT Bench} \\
\midrule
LLaMA-2-7B & 3.97 \\
LLaMA-2-7B (alpaca-gpt4) & 4.31 \\
LLaMA-2-7B (self-instruct--Teacher: Mistral) & 4.96 \\
LLaMA-2-7B-chat Official implementation & 6.27 \\
LLaMA-2-7B (FANNO--Teacher: LLaMA-2-Chat) & 4.65 \\
LLaMA-2-7B (FANNO--Teacher: Mistral) & 5.11 \\
\bottomrule
\end{tabular}
}
\caption{Comparison of Different Models on MT Bench}
\label{tab:mt_bench}
\vspace{-10pt}
\end{table}

\subsection{Results}

The comparative experiments of \mname with other models demonstrate the superiority of our work. 
(1) For diverse base models like LLaMA and Mistral, our framework consistently achieves top rankings in the LLM-open-leaderboard, even rivaling the models fine-tuned with Alpaca-GPT4-Cleaned, which underwent augmentation with proprietary models and manual selection. (2) Compared to other similar automatic instruction annotation frameworks like Humpback,
Muffin, WizardLM, we adhered to the principle of fairness as much as possible by fully utilizing the officially published datasets, and experiments proved that \mname achieved excellent results with a smaller dataset. 

Using MT-Bench in Table~\ref{tab:mt_bench}, we observe that our model outperforms those fine-tuned using Alpaca-GPT4-clean, highlighting the effectiveness of \mname. Moreover, compared with \textit{self-instruct}, we relieve the need for manually labeled data and achieve better performance, while naive \textit{self-instruct} with Mistral does not yield optimal results.
However, it is understandably inferior to the LLaMA2-7B-Chat model, which benefits from extensive fine-tuning and RLHF alignment.

\begin{table*}[h]
\centering
\scalebox{0.80}{
  \centering
  \begin{tabular}{lcrrrr}
    \toprule
    \textbf{Model} & \textbf{ARC} & \textbf{HellaSwag} & \textbf{MMLU} & \textbf{TruthfulQA} & \textbf{Average}\\
    \midrule
    \textrm{LLaMA-2-Base} &  53.07 & 78.59 & 46.87 & 38.76 & 54.32 \\
    \textrm{LLaMA-2-Chat} &  52.90 & 78.55 & \cellcolor{myred!25}48.32 & 45.57 & 56.34 \\
    \textrm{Vicuna-7b-v1.3} &  50.43 & 76.92 & \cellcolor{mygreen!25}48.14 & 47.01 & 55.62 \\
    Humback $M_0$ &  \cellcolor{myred!25}56.31 & \cellcolor{myred!25} 81.20 & 47.45 & 47.59 & \cellcolor{mygreen!25}58.13 \\
    Humback $M_1$ &  52.99 & 78.57 & 45.48 & 41.45 & 54.65 \\
    \textrm{WizardLM-7b} &  54.95 & 77.85 & 45.79 & \cellcolor{mygreen!25}48.29 & 56.72 \\
    \mname &  \cellcolor{mygreen!25}55.46 & \cellcolor{mygreen!25}79.29 & 46.58 & \cellcolor{myred!25}52.05 & \cellcolor{myred!25}58.35 \\
    \bottomrule
    \end{tabular}
   }
  \caption{Benchmark results evaluated by the official Huggingface Open LLM Leaderboard platform.}
  \label{tab:openllm2}
  \vspace{-6pt}
\end{table*}

Models refined through \mname exhibit notable enhancements in the TruthfulQA metric and show measurable improvements across three other metrics. This advancement is attributed to the integration of supplementary information via the RAG component and self-reflective teacher model, thereby improving the model's proficiency in delivering more faithful outputs and bolstering TruthfulQA scores. Slight improvements in ARC, HellaSwag, MMLU metrics are credited to the elevated challenge and diversity of the instructions, as depicted in Table~\ref{tab:openllm1}. We also uploaded our model to Huggingface Open LLM Leaderboard and compared our results with models like Vicuna, and Humpback, which are shown in Table~\ref{tab:openllm2}. As shown in Figure~\ref{fig:alpaca_eval_result}, our model marginally outperforms the Alpaca-GPT4-Cleaned's fine-tuned variant on the AlpacaEval benchmark, attesting to the superiority of our \mname framework. 

\begin{figure}[ht]
    \begin{minipage}[b]{0.45\textwidth}
        \centering
        \resizebox{0.95\linewidth}{!}{\includegraphics[width=1.1\textwidth]{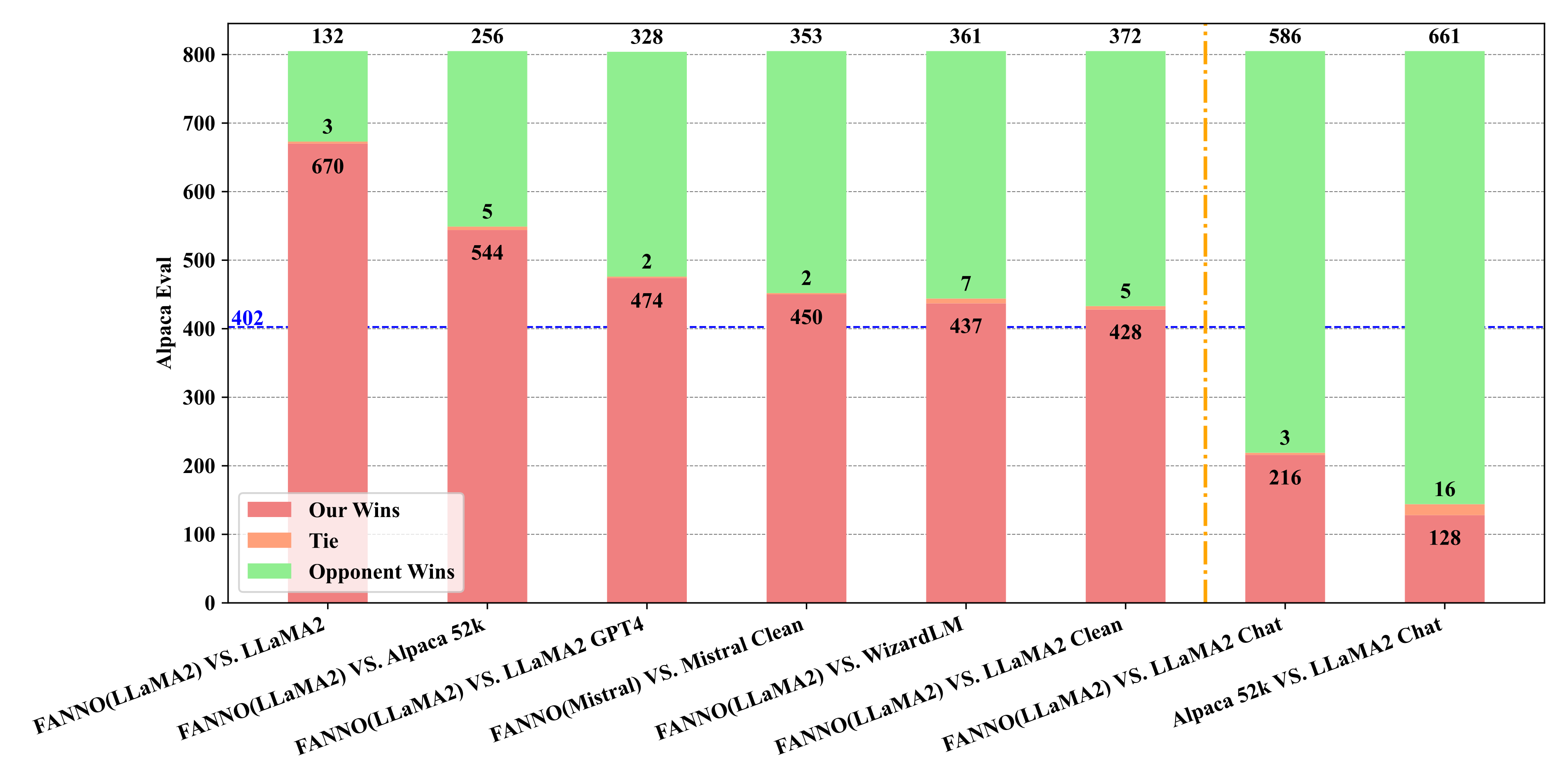}}
        \caption{AlpacaEval Result}
        \label{fig:fanno_diversity_comparison}
    \end{minipage}
    \vspace{-10pt}
\end{figure}

\begin{table*}[!htbh]
\centering
\scalebox{0.80}{
\begin{tabular}{l|l|cccc|l}
\hline
 \multirow{2}{*}{\textbf{ID}} & \multirow{2}{*}{\textbf{Configuration}} & \multicolumn{4}{c|}{\textbf{Open LLM Leaderboard}} & \multirow{2}{*}{\textbf{Average}} \\
 &  & \textbf{ARC} & \textbf{HellaSwag} & \textbf{MMLU} & \textbf{TruthfulQA} &    \\
\hline
0&Base       & 54.44 & 78.66 & 44.69 & 46.02 & 55.95 \\
1& Pre-Screen    & 55.46 & 78.51 & 46.00 & 45.85 & 56.44 \\
\hline
2&\mname (w/o Iter and UCB) & 54.69 & 79.18  & 45.92 & 50.19  & 57.50  \\
3&\mname (w/o UCB) & 55.63 & 79.43 & 44.84 & 51.16 & 57.77 \\
\hline
4&\mname (w/ OD) & 55.46 & 78.31 & 44.99 & 45.68 & 56.11 \\
5&\mname (w/ RAG) & 55.29 & 78.41 & 45.80 & 45.37  & 56.22\\
6&\mname (w/ RAG+) & 55.03 & 78.46 & 47.02 & 46.26  & 56.69\\
7&\mname  & 55.63  & 79.45 & 46.84 & 51.01 & 58.23  \\ 
\hline
\end{tabular}
\begin{tikzpicture}[ remember picture, overlay]
    \node[rotate = 270] at (-14.86, 1.12) {$\underbrace{\hspace{0.75cm}}$};
    \node at (-15.16,1.08) {\textcircled{\tiny 1}};
    \node[rotate = 270] at (-14.86, 0.09) {$\underbrace{\hspace{0.75cm}}$};
    \node at (-15.16,0.0) {\textcircled{\tiny 2}};
    \node[rotate = 270] at (-14.86, -1.27) {$\underbrace{\hspace{1.7cm}}$};
    \node at (-15.16,-1.35) {\textcircled{\tiny 3}};
\end{tikzpicture}
}
\caption{
Ablation results from the lm-evaluation-harness\citep{eval-harness}. (0). Basic framework: simply generate instructions by documents and generate responses by instructions without any optimization. (1). Add Pre-Screen module into the basic framework. (2). \mname without Iteration and UCB-selection. (3). \mname without UCB-selection. (4). \mname with the original document. (5). \mname with RAG module.
(6). \mname with RAG module and supplementary materials. 
(7). The complete version of \mname.
}
\label{tb:ablation}
\vspace{-4.2mm}
\end{table*}

\subsection{Ablation Study} 
To enhance our understanding of the functionality of each module within \textit{\mname}, we undertook ablation studies on its four components, as delineated in Table~\ref{tb:ablation}. Our findings reveal:

\begin{itemize}[label=\textbullet, itemsep=0pt, parsep=0pt, topsep=2pt, leftmargin=10pt]
\item \textbf{Orthogonality of Components and Separate Optimization}
We replaced each component with a random strategy, and the experiments show that each module positively affects the model's performance, and using more advanced strategies yields better results. \textcircled{\tiny 1} indicated that the Pre-Screen strategy helps to enhance the quality and thematic diversity of the raw documents. Configurations (3) and (7) demonstrated that using the UCB strategy in instruction augmentation balances the complexity and diversity of the generations, achieving higher diversity compared to random sampling. 
The notable growth in the MMLU result, as indicated by the \textcircled{\tiny 2} \& 7 combinations, revealed that iterative enhancements in conjunction with the UCB strategy were paramount. The UCB's proactive selection of high-quality data for augmentation facilitates a gradual evolution towards more effective methodologies as the iteration progresses.

\item \textbf{Generalizing Boosting Diversity and Complexity}
As introduced in Section~\ref{framework}, \mname uses randomness, deduplication, and carefully designed prompts to increase the diversity of themes and tasks in instructions as much as possible. In this way, \mname tends to break away from reliance on the corresponding unlabeled documents, enhancing the generalizability of instructions. Ablation experiments \textcircled{\tiny 1} \& 7 proved that texts with higher generalizability exhibit more diversity and complexity, which is more beneficial for activating the capabilities of the base models.

\item \textbf{Knowledge Supplementation Promotes Instruction Quality}
The results of \textcircled{\tiny 3} demonstrated that it is necessary to incorporate RAG or supplement knowledge with the help of a teacher model is necessary. 
We discovered that considering only the direct generation of the teacher model yielded the best results compared to the document-based response, particularly on TruthfulQA. This indicates that instructions generated with the \mname framework are more general and less reliable on the corresponding document.
We used RAG in an experiment to pinpoint the most relevant content for the instruction, and experiments 5 \& 6 support our assertion that more is better. 
Other discussions about the truthfulness are covered in Section~\ref{analyses}. RAG+ used larger datasets than RAG, but both were from Wikipedia.\footnote{The RAG used 2.73GB of data from Wikipedia's introduction section, and the RAG+ used 20.28GB of data from Wikipedia.}
\end{itemize}

\section{Analyses} \label{analyses}

In this section, we discuss how diversity, correctness, and complexity are promoted in each stage.

\subsection{Analyses of the Augmented Instruction Data} 
We analyze and illustrate the generated instructions of our dataset from 4 aspects:

\paragraph{Length} To study the distribution of the length of instructions, we tokenize each instruction combined with input and count the words within it as its length. Figure~\ref{fig:fanno_instruction_length_distribution} and Figure~\ref{fig:alpaca_instruction_length_distribution} in Appendix~\ref{appendix:quality_length_diversity} illustrate the distribution of instruction length for \mname and Alpaca-Cleaned, respectively. The results show that \mname instructions are more balanced than Alpaca-Cleaned and the mean value of lengths is higher than that of Alpaca, which indicates a better performance.

\paragraph{Diversity} Inspired by \textsc{Self-Instruct} \citep{wang2022self}, the verb-noun pairs in instructions to represent the types and tasks of instructions are identified and extracted, which exhibits diversity. As Figure~\ref{fig:fanno_instruction_verb_noun} and Figure~\ref{fig:alpaca_clean_instruction_verb_noun} in Appendix~\ref{appendix:quality_length_diversity} depicted, \mname instructions possess more challenging verb-noun pairs than Alpaca-Cleaned, which indicates more challenging tasks. The extraction is completed by Berkeley Neural Parser \citep{kitaev-klein-2018-constituency,kitaev-etal-2019-multilingual}.

\paragraph{Quality and Complexity} To evaluate the quality and complexity of instruction-response pairs, we utilize Deita-quality-scorer model and Deita-complexity-scorer model~\citep{liu2023what} as an evaluator to score our instructions. Figure~\ref{fig:quality_complexity} in Appendix~\ref{appendix:quality_length_diversity} shows the quality and complexity comparison between \mname and Alpaca-Cleaned, of which the result shows that \mname instructions possess a more balanced complexity distribution and higher average quality. The corresponding prompts can be found in Table~\ref{table:text_quality_scorer_prompt} and Table~\ref{table:text_complexity_scorer_prompt} in the appendix.



\subsection{Randomness Tag Boosting the Complexity} \label{sec:complex_discuss}


\begin{table}[!t]
\centering
\footnotesize
\begin{tabular}{@{}lcc@{}}
\toprule
\textbf{Score} & \textbf{Tagged} & \textbf{Untagged} \\ \midrule
0 (bad) & 18 & 24 \\ 
1 (everyday) & 39 & 78 \\ 
2 (expert) & 143 & 98 \\  \bottomrule
\end{tabular}
\caption{Randomness Tag Evaluation Results}
\label{tab:human_test_results}
\vspace{-10pt}
\end{table}

Randomness tags serve as additional requirements for the teacher model when generating instructions, enhancing their complexity. To demonstrate the effectiveness of this strategy, generated instructions are manually annotated to evaluate their complexity, as discussed in Section~\ref{sec:evaluation}. We randomly sampled 200 instances from two datasets for manual testing: one utilizing the Random tag strategy (Tagged) and the other generated directly (Untagged). Results are summarized in Table~\ref{tab:human_test_results}, with example instructions detailed in Appendix~\ref{appendix:human_eval_complexity}. From the results, it is evident that instructions with random tags exhibit a significant increase in complexity, manifested by a greater number of expert-level instructions and fewer daily instructions. It is worth noting that instructions with random tags exhibit a tendency towards greater length, including few overly complex tasks that are difficult to answer (classified as expert-level instructions), which indicates a potential need for further refinement.

\begin{figure}[ht]
    \begin{minipage}[b]{0.45\textwidth}
        \centering
        \includegraphics[width=1.1\textwidth]{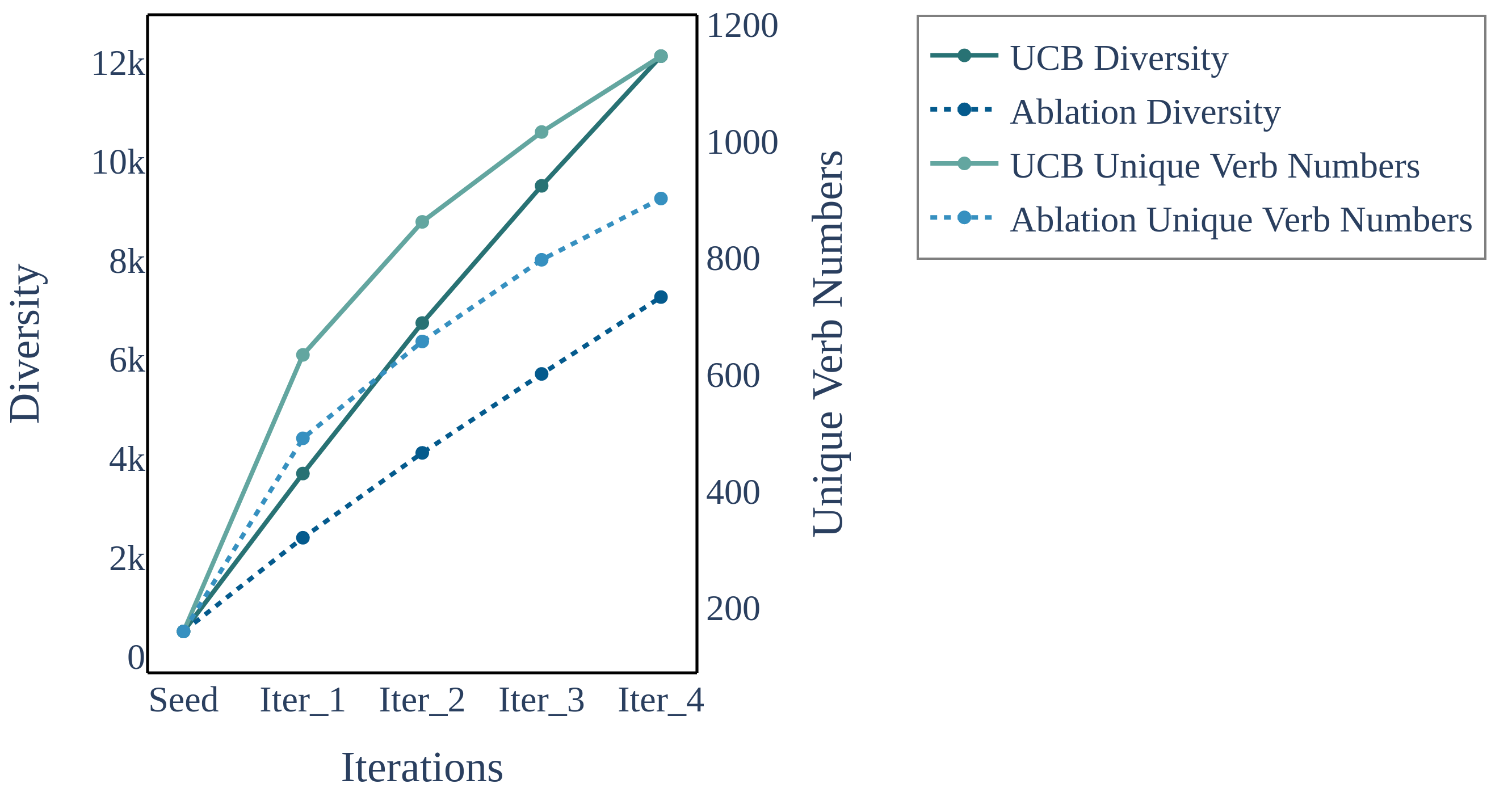}
        \caption{The verbs-noun statistics data grows with iteration}
        \label{fig:fanno_diversity_comparison}
    \end{minipage}
    \hfill
    \begin{minipage}[b]{0.45\textwidth}
        \centering
        \includegraphics[width=1.1\textwidth]{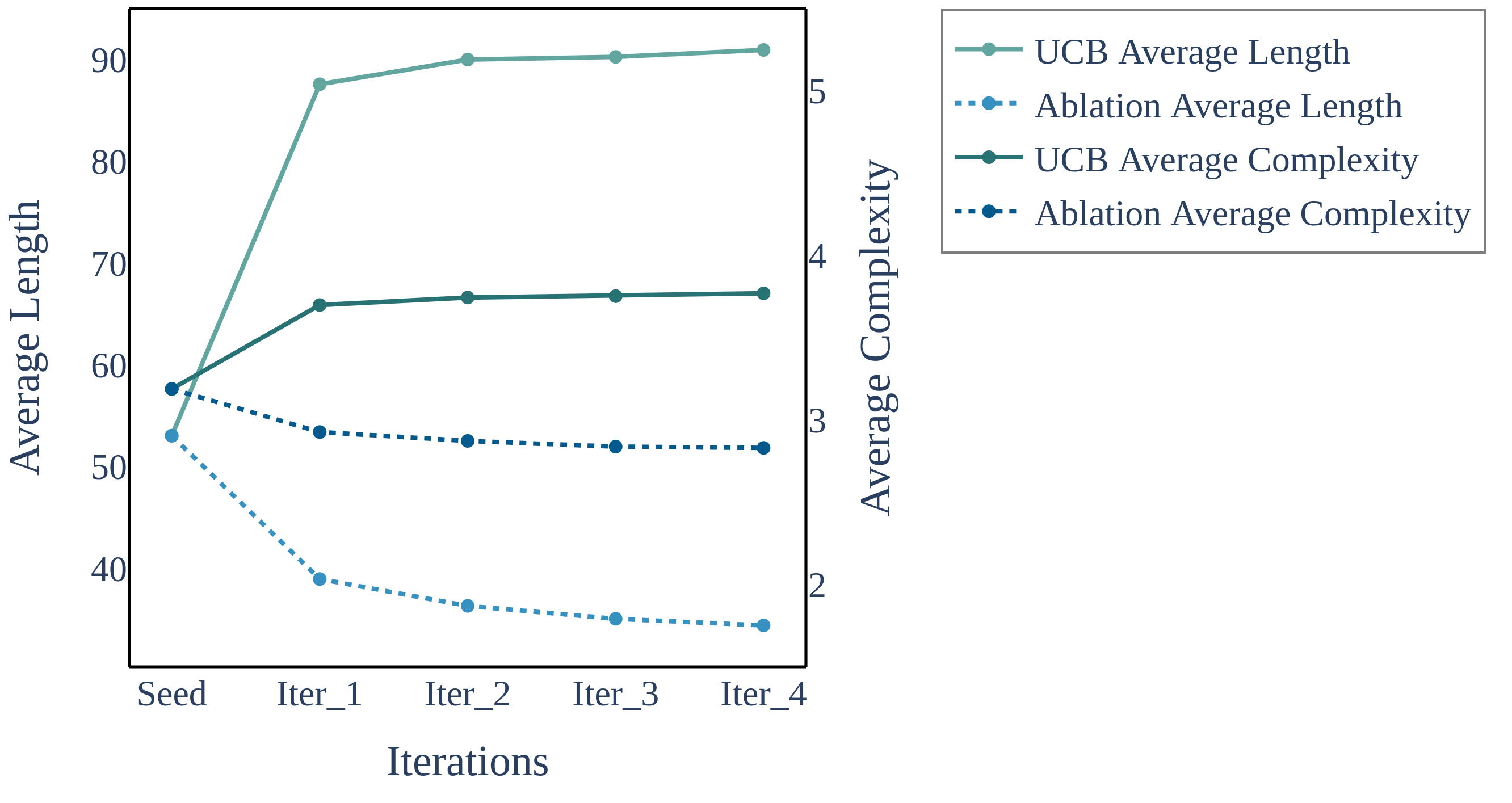}
        \caption{The instruction length (complexity) grows with iteration }
        \label{fig:fanno_length_comparison}
    \end{minipage}
    \vspace{-10pt}
\end{figure}


\begin{table*}[h!]
\centering
\scalebox{0.90}{
\begin{tabular}{l|cccc|l}
\hline
& \multicolumn{4}{c|}{\textbf{Open LLM Leaderboard}} &  \multirow{2}{*}{\textbf{Average}} \\
& \textbf{ARC} & \textbf{HellaSwag} & \textbf{MMLU} & \textbf{TruthfulQA} &   \\
\hline
Direct Responses & 54.86 & 79.33 & 45.56 & 47.40 & 56.79   \\
Cautious Response & 54.44 & 79.17 & 45.88 & 47.29 & 56.69   \\
\hline
Faithful Responses & 55.03 & 78.63 & 45.66 & 42.82 & 55.54   \\
Adaptive Responses & 55.63 & 78.71 & 45.86 & 42.76 & 55.74   \\
\hline
\end{tabular}
}
\begin{tikzpicture}[ remember picture, overlay]
    \node[rotate = 270] at (-11.70, 0.09) {$\underbrace{\hspace{0.68cm}}$};
    \node at (-12.82,0.07) {direct-based};
    \node[rotate = 270] at (-11.70, -0.7) {$\underbrace{\hspace{0.68cm}}$};
    \node at (-12.87,-0.7) {doc-based};
\end{tikzpicture}
\caption{
Comparison results of four types response on the Open LLM 
}
\label{tb:type_ablation}
\vspace{-4.2mm}
\end{table*}

\subsection{UCB Bootstrap Iteration Improve Instruction Complexity while Maintaining the Diversity}



UCB Bootstrap is employed to actively stabilize the process of instruction improvement. For illustration, we monitored the diversity and complexity of instructions during iterations and compared it with a random selection strategy. Note that to simplify the process, we use the length of instructions in words as the measure of quality. As depicted in Figure~\ref{fig:fanno_diversity_comparison} and Figure~\ref{fig:fanno_length_comparison}, we observed an increase in both the average complexity and diversity scores as the iteration progressed, consistent with our expectations. We analyze that UCB prioritizes exploring longer instructions for few-shot instruction generation, resulting in more challenging instructions. Additionally, UCB exhibits a preference for selecting newly generated instructions, as novel few-shot combinations tend to ignite the model's creativity.

\subsection{Truthfulness is Less Important for Capability Activation}  \label{Truthfulness}

Previous work~\citep{zhou2023lima, liu2023deita} has explored the diversity, complexity, and fidelity of instructions that enhance large models' capabilities. We further investigate the truthfulness of responses to instructions, noting that responses often seem accurate but contain illusions and fabricated information, potentially affecting instruction fine-tuning. 

To address this, we selected about 1,000 expert-level instructions from the \mname dataset, then prompted LLM to generate four different responses for each instruction with the following settings: 
\begin{enumerate}[label=\textbullet, itemsep=2pt, parsep=0pt, topsep=2pt, leftmargin=*]
    \item Direct Response: Models provide answers regardless of correctness.
    \item Cautious Response: Models may acknowledge a lack of knowledge.
    \item Faithful Response: Models generate answers solely based on provided documents.
    \item Adaptive Response: Models use relevant information from provided documents to generate answers.
\end{enumerate}

From the result in Table~\ref{tb:type_ablation}, an intriguing finding is that direct responses from the model, which contains a substantial presence of illusions, outperformed document-based ones, particularly in the TruthfulQA task. This might suggest that providing human-like and consistent responses, even with false data, can also improve the model's capabilities during SFT.
We also need to point out \mname also introduces external sources of information such as the knowledge of the teacher model itself, which likewise results in some illusory responses.

\section{Limitations}
While \mname has demonstrated outstanding performance, several limitations must be acknowledged.
The responses are not entirely dependent on the document, leading to the introduction of certain hallucinations in the fine-tuning data, as discussed in Section ~\ref{Truthfulness}. This suggests that the model's reliance on the provided context needs to be strengthened to improve factual consistency.
The simplistic approach of equating instruction length with its value is rather crude. The true value of an instruction is influenced by various factors such as difficulty, quality, and novelty. Future work will aim to develop a more nuanced understanding and evaluation of instruction value. 
The quality of generated instructions is contingent upon the capabilities of both the generator and the evaluator. This process is sensitive to the teacher model and the prompts used, indicating a need for designing prompts that are specifically tailored to the model.
Addressing these limitations will be a focus of our future work.

\section{Conclusion}
The development of instruction tuning datasets has been hindered by the high cost and labor-intensive nature. In this paper, we introduced \mname, an autonomous and low-cost framework that addresses these challenges by streamlining the annotation process with open-sourced LLMs. \mname efficiently generates datasets of high quality, diversity, and complexity through a structured process involving pre-screening, instruction generation, and response generation. This unified process eliminates the need for pre-existing annotated data or costly API calls, advancing the instruction data development. Empirical experiments validate the efficacy of \mname, underscoring the framework's potential to democratize access to high-quality instruction datasets. \mname enables access to top-quality datasets with reduced cost and effort, driving progress in LLM applications.
%

\clearpage
\newpage 



\bibliography{anthology,custom}

\begin{thebibliography}{51}
\providecommand{\natexlab}[1]{#1}

\bibitem[{Beeching et~al.(2023)Beeching, Fourrier, Habib, Han, Lambert, Rajani, Sanseviero, Tunstall, and Wolf}]{open-llm-leaderboard}
Edward Beeching, Clémentine Fourrier, Nathan Habib, Sheon Han, Nathan Lambert, Nazneen Rajani, Omar Sanseviero, Lewis Tunstall, and Thomas Wolf. 2023.
\newblock Open llm leaderboard.
\newblock \url{https://huggingface.co/spaces/HuggingFaceH4/open_llm_leaderboard}.

\bibitem[{Chiang et~al.(2023)Chiang, Li, Lin, Sheng, Wu, Zhang, Zheng, Zhuang, Zhuang, Gonzalez, Stoica, and Xing}]{vicuna2023}
Wei-Lin Chiang, Zhuohan Li, Zi~Lin, Ying Sheng, Zhanghao Wu, Hao Zhang, Lianmin Zheng, Siyuan Zhuang, Yonghao Zhuang, Joseph~E. Gonzalez, Ion Stoica, and Eric~P. Xing. 2023.
\newblock \href {https://lmsys.org/blog/2023-03-30-vicuna/} {Vicuna: An open-source chatbot impressing gpt-4 with 90\%* chatgpt quality}.

\bibitem[{Chung et~al.(2022)Chung, Hou, Longpre, Zoph, Tay, Fedus, Li, Wang, Dehghani, Brahma et~al.}]{chung2022scaling}
Hyung~Won Chung, Le~Hou, Shayne Longpre, Barret Zoph, Yi~Tay, William Fedus, Eric Li, Xuezhi Wang, Mostafa Dehghani, Siddhartha Brahma, et~al. 2022.
\newblock Scaling {I}nstruction-{F}inetuned {L}anguage {M}odels.
\newblock \emph{arXiv preprint arXiv:2210.11416}.

\bibitem[{Clark et~al.(2018)Clark, Cowhey, Etzioni, Khot, Sabharwal, Schoenick, and Tafjord}]{clark2018think}
Peter Clark, Isaac Cowhey, Oren Etzioni, Tushar Khot, Ashish Sabharwal, Carissa Schoenick, and Oyvind Tafjord. 2018.
\newblock \href {https://arxiv.org/abs/1803.05457} {Think you have solved question answering? try arc, the ai2 reasoning challenge}.
\newblock \emph{Preprint}, arXiv:1803.05457.

\bibitem[{Conover et~al.(2023)Conover, Hayes, Mathur, Meng, Xie, Wan, Ghodsi, Wendell, and Zaharia}]{dolly2023}
Mike Conover, Matt Hayes, Matt Mathur, Xiangrui Meng, Jianwei Xie, Jun Wan, Ali Ghodsi, Patrick Wendell, and Patrick Zaharia. 2023.
\newblock \href {https://github.com/databrickslabs/dolly} {Hello dolly: Democratizing the magic of chatgpt with open models.}

\bibitem[{Du et~al.(2023)Du, Zong, and Zhang}]{du2023mods}
Qianlong Du, Chengqing Zong, and Jiajun Zhang. 2023.
\newblock \href {https://arxiv.org/abs/2311.15653} {Mods: Model-oriented data selection for instruction tuning}.
\newblock \emph{Preprint}, arXiv:2311.15653.

\bibitem[{Fan et~al.(2023)Fan, Gokkaya, Harman, Lyubarskiy, Sengupta, Yoo, and Zhang}]{fan2023large}
Angela Fan, Beliz Gokkaya, Mark Harman, Mitya Lyubarskiy, Shubho Sengupta, Shin Yoo, and Jie~M. Zhang. 2023.
\newblock \href {https://arxiv.org/abs/2310.03533} {Large language models for software engineering: Survey and open problems}.
\newblock \emph{Preprint}, arXiv:2310.03533.

\bibitem[{Gao et~al.(2023)Gao, Tow, Abbasi, Biderman, Black, DiPofi, Foster, Golding, Hsu, Le~Noac'h, Li, McDonell, Muennighoff, Ociepa, Phang, Reynolds, Schoelkopf, Skowron, Sutawika, Tang, Thite, Wang, Wang, and Zou}]{eval-harness}
Leo Gao, Jonathan Tow, Baber Abbasi, Stella Biderman, Sid Black, Anthony DiPofi, Charles Foster, Laurence Golding, Jeffrey Hsu, Alain Le~Noac'h, Haonan Li, Kyle McDonell, Niklas Muennighoff, Chris Ociepa, Jason Phang, Laria Reynolds, Hailey Schoelkopf, Aviya Skowron, Lintang Sutawika, Eric Tang, Anish Thite, Ben Wang, Kevin Wang, and Andy Zou. 2023.
\newblock \href {https://doi.org/10.5281/zenodo.10256836} {A framework for few-shot language model evaluation}.

\bibitem[{Geng et~al.(2023)Geng, Gudibande, Liu, Wallace, Abbeel, Levine, and Song}]{koala_blogpost_2023}
Xinyang Geng, Arnav Gudibande, Hao Liu, Eric Wallace, Pieter Abbeel, Sergey Levine, and Dawn Song. 2023.
\newblock \href {https://bair.berkeley.edu/blog/2023/04/03/koala/} {Koala: A dialogue model for academic research}.
\newblock Blog post.

\bibitem[{Hendrycks et~al.(2021)Hendrycks, Burns, Basart, Zou, Mazeika, Song, and Steinhardt}]{hendrycks2021measuring}
Dan Hendrycks, Collin Burns, Steven Basart, Andy Zou, Mantas Mazeika, Dawn Song, and Jacob Steinhardt. 2021.
\newblock \href {https://arxiv.org/abs/2009.03300} {Measuring massive multitask language understanding}.
\newblock \emph{Preprint}, arXiv:2009.03300.

\bibitem[{Hu et~al.(2021)Hu, Shen, Wallis, Allen-Zhu, Li, Wang, Wang, and Chen}]{hu2021lora}
Edward~J. Hu, Yelong Shen, Phillip Wallis, Zeyuan Allen-Zhu, Yuanzhi Li, Shean Wang, Lu~Wang, and Weizhu Chen. 2021.
\newblock \href {https://arxiv.org/abs/2106.09685} {Lora: Low-rank adaptation of large language models}.
\newblock \emph{Preprint}, arXiv:2106.09685.

\bibitem[{Jiang et~al.(2023{\natexlab{a}})Jiang, Sablayrolles, Mensch, Bamford, Chaplot, de~las Casas, Bressand, Lengyel, Lample, Saulnier, Lavaud, Lachaux, Stock, Scao, Lavril, Wang, Lacroix, and Sayed}]{jiang2023mistral}
Albert~Q. Jiang, Alexandre Sablayrolles, Arthur Mensch, Chris Bamford, Devendra~Singh Chaplot, Diego de~las Casas, Florian Bressand, Gianna Lengyel, Guillaume Lample, Lucile Saulnier, Lélio~Renard Lavaud, Marie-Anne Lachaux, Pierre Stock, Teven~Le Scao, Thibaut Lavril, Thomas Wang, Timothée Lacroix, and William~El Sayed. 2023{\natexlab{a}}.
\newblock \href {https://arxiv.org/abs/2310.06825} {Mistral 7b}.
\newblock \emph{Preprint}, arXiv:2310.06825.

\bibitem[{Jiang et~al.(2023{\natexlab{b}})Jiang, Li, Zhang, Huang, Lin, and Chen}]{jiang2023tigerscore}
Dongfu Jiang, Yishan Li, Ge~Zhang, Wenhao Huang, Bill~Yuchen Lin, and Wenhu Chen. 2023{\natexlab{b}}.
\newblock \href {https://arxiv.org/abs/2310.00752} {Tigerscore: Towards building explainable metric for all text generation tasks}.
\newblock \emph{Preprint}, arXiv:2310.00752.

\bibitem[{Kitaev et~al.(2019)Kitaev, Cao, and Klein}]{kitaev-etal-2019-multilingual}
Nikita Kitaev, Steven Cao, and Dan Klein. 2019.
\newblock \href {https://doi.org/10.18653/v1/P19-1340} {Multilingual constituency parsing with self-attention and pre-training}.
\newblock In \emph{Proceedings of the 57th Annual Meeting of the Association for Computational Linguistics}, pages 3499--3505, Florence, Italy. Association for Computational Linguistics.

\bibitem[{Kitaev and Klein(2018)}]{kitaev-klein-2018-constituency}
Nikita Kitaev and Dan Klein. 2018.
\newblock \href {https://doi.org/10.18653/v1/P18-1249} {Constituency parsing with a self-attentive encoder}.
\newblock In \emph{Proceedings of the 56th Annual Meeting of the Association for Computational Linguistics (Volume 1: Long Papers)}, pages 2676--2686, Melbourne, Australia. Association for Computational Linguistics.

\bibitem[{Li et~al.(2023{\natexlab{a}})Li, Chen, Chen, He, Huang, Gu, and Zhou}]{Li2023ReflectionTuningDR}
Ming Li, Lichang Chen, Jiuhai Chen, Shwai He, Heng Huang, Jiuxiang Gu, and Tianyi Zhou. 2023{\natexlab{a}}.
\newblock \href {https://api.semanticscholar.org/CorpusID:264288970} {Reflection-tuning: Data recycling improves llm instruction-tuning}.
\newblock \emph{ArXiv}, abs/2310.11716.

\bibitem[{Li et~al.(2024)Li, Yu, Zhou, Schick, Levy, Zettlemoyer, Weston, and Lewis}]{li2024selfalignment}
Xian Li, Ping Yu, Chunting Zhou, Timo Schick, Omer Levy, Luke Zettlemoyer, Jason Weston, and Mike Lewis. 2024.
\newblock \href {https://arxiv.org/abs/2308.06259} {Self-alignment with instruction backtranslation}.
\newblock \emph{Preprint}, arXiv:2308.06259.

\bibitem[{Li et~al.(2023{\natexlab{b}})Li, Zhang, Dubois, Taori, Gulrajani, Guestrin, Liang, and Hashimoto}]{alpaca_eval}
Xuechen Li, Tianyi Zhang, Yann Dubois, Rohan Taori, Ishaan Gulrajani, Carlos Guestrin, Percy Liang, and Tatsunori~B. Hashimoto. 2023{\natexlab{b}}.
\newblock Alpacaeval: An automatic evaluator of instruction-following models.
\newblock \url{https://github.com/tatsu-lab/alpaca_eval}.

\bibitem[{Lin et~al.(2022)Lin, Hilton, and Evans}]{lin2022truthfulqa}
Stephanie Lin, Jacob Hilton, and Owain Evans. 2022.
\newblock \href {https://arxiv.org/abs/2109.07958} {Truthfulqa: Measuring how models mimic human falsehoods}.
\newblock \emph{Preprint}, arXiv:2109.07958.

\bibitem[{Liu et~al.(2023{\natexlab{a}})Liu, Zeng, He, Jiang, and He}]{liu2023what}
Wei Liu, Weihao Zeng, Keqing He, Yong Jiang, and Junxian He. 2023{\natexlab{a}}.
\newblock \href {https://arxiv.org/abs/2312.15685} {What makes good data for alignment? a comprehensive study of automatic data selection in instruction tuning}.
\newblock \emph{Preprint}, arXiv:2312.15685.

\bibitem[{Liu et~al.(2023{\natexlab{b}})Liu, Zeng, He, Jiang, and He}]{liu2023deita}
Wei Liu, Weihao Zeng, Keqing He, Yong Jiang, and Junxian He. 2023{\natexlab{b}}.
\newblock \href {https://arxiv.org/abs/2312.15685} {What makes good data for alignment? a comprehensive study of automatic data selection in instruction tuning}.
\newblock \emph{Preprint}, arXiv:2312.15685.

\bibitem[{Lou et~al.(2024)Lou, Zhang, Xie, Sun, Ahn, Xu, Su, and Yin}]{lou2024muffin}
Renze Lou, Kai Zhang, Jian Xie, Yuxuan Sun, Janice Ahn, Hanzi Xu, Yu~Su, and Wenpeng Yin. 2024.
\newblock \href {https://openreview.net/forum?id=1vrS1zwekw} {{MUFFIN}: Curating multi-faceted instructions for improving instruction following}.
\newblock In \emph{The Twelfth International Conference on Learning Representations}.

\bibitem[{Lou et~al.(2023)Lou, Zhang, and Yin}]{lou2023prompt}
Renze Lou, Kai Zhang, and Wenpeng Yin. 2023.
\newblock Is prompt all you need? no. a comprehensive and broader view of instruction learning.
\newblock \emph{arXiv preprint arXiv:2303.10475}.

\bibitem[{Mishra et~al.(2022)Mishra, Khashabi, Baral, and Hajishirzi}]{mishra2022cross}
Swaroop Mishra, Daniel Khashabi, Chitta Baral, and Hannaneh Hajishirzi. 2022.
\newblock {C}ross-{T}ask {G}eneralization via {N}atural {L}anguage {C}rowdsourcing {I}nstructions.
\newblock In \emph{Proceedings of the 60th Annual Meeting of the Association for Computational Linguistics (Volume 1: Long Papers)}, pages 3470--3487.

\bibitem[{Mitra et~al.(2024)Mitra, Khanpour, Rosset, and Awadallah}]{mitra2024orcamath}
Arindam Mitra, Hamed Khanpour, Corby Rosset, and Ahmed Awadallah. 2024.
\newblock \href {https://arxiv.org/abs/2402.14830} {Orca-math: Unlocking the potential of slms in grade school math}.
\newblock \emph{Preprint}, arXiv:2402.14830.

\bibitem[{OpenAI(2023)}]{openai2023gpt4}
OpenAI. 2023.
\newblock \href {https://arxiv.org/abs/2303.08774} {Gpt-4 technical report}.
\newblock \emph{Preprint}, arXiv:2303.08774.

\bibitem[{Ouyang et~al.(2022)Ouyang, Wu, Jiang, Almeida, Wainwright, Mishkin, Zhang, Agarwal, Slama, Ray et~al.}]{ouyang2022training}
Long Ouyang, Jeff Wu, Xu~Jiang, Diogo Almeida, Carroll~L Wainwright, Pamela Mishkin, Chong Zhang, Sandhini Agarwal, Katarina Slama, Alex Ray, et~al. 2022.
\newblock Training language models to follow instructions with human feedback.
\newblock \emph{arXiv preprint arXiv:2203.02155}.

\bibitem[{Penedo et~al.(2023)Penedo, Malartic, Hesslow, Cojocaru, Cappelli, Alobeidli, Pannier, Almazrouei, and Launay}]{penedo2023refinedweb}
Guilherme Penedo, Quentin Malartic, Daniel Hesslow, Ruxandra Cojocaru, Alessandro Cappelli, Hamza Alobeidli, Baptiste Pannier, Ebtesam Almazrouei, and Julien Launay. 2023.
\newblock \href {https://arxiv.org/abs/2306.01116} {The refinedweb dataset for falcon llm: Outperforming curated corpora with web data, and web data only}.
\newblock \emph{Preprint}, arXiv:2306.01116.

\bibitem[{Peng et~al.(2023)Peng, Li, He, Galley, and Gao}]{peng2023gpt4llm}
Baolin Peng, Chunyuan Li, Pengcheng He, Michel Galley, and Jianfeng Gao. 2023.
\newblock Instruction tuning with gpt-4.
\newblock \emph{arXiv preprint arXiv:2304.03277}.

\bibitem[{Press et~al.(2023)Press, Zhang, Min, Schmidt, Smith, and Lewis}]{press2023measuring}
Ofir Press, Muru Zhang, Sewon Min, Ludwig Schmidt, Noah~A. Smith, and Mike Lewis. 2023.
\newblock \href {https://arxiv.org/abs/2210.03350} {Measuring and narrowing the compositionality gap in language models}.
\newblock \emph{Preprint}, arXiv:2210.03350.

\bibitem[{Raschka(2023)}]{raschka_lora_2023}
Sebastian Raschka. 2023.
\newblock \href {[https://lightning.ai/pages/community/lora-insights/](https://lightning.ai/pages/community/lora-insights/)} {Finetuning llms with lora and qlora: Insights from hundreds of experiments}.
\newblock \emph{Lightning AI}.

\bibitem[{Reimers and Gurevych(2019)}]{reimers-2019-sentence-bert}
Nils Reimers and Iryna Gurevych. 2019.
\newblock \href {http://arxiv.org/abs/1908.10084} {Sentence-bert: Sentence embeddings using siamese bert-networks}.
\newblock In \emph{Proceedings of the 2019 Conference on Empirical Methods in Natural Language Processing}. Association for Computational Linguistics.

\bibitem[{Robbins and Monro(1951)}]{robbins1951stochastic}
Herbert Robbins and Sutton Monro. 1951.
\newblock A stochastic approximation method.
\newblock \emph{The annals of mathematical statistics}, pages 400--407.

\bibitem[{Sanh et~al.(2022)Sanh, Webson, Raffel, Bach, Sutawika, Alyafeai, Chaffin, Stiegler, Raja, Dey et~al.}]{sanh2021multitask}
Victor Sanh, Albert Webson, Colin Raffel, Stephen Bach, Lintang Sutawika, Zaid Alyafeai, Antoine Chaffin, Arnaud Stiegler, Arun Raja, Manan Dey, et~al. 2022.
\newblock Multitask {P}rompted {T}raining {E}nables {Z}ero-{S}hot {T}ask {G}eneralization.
\newblock In \emph{International Conference on Learning Representations}.

\bibitem[{Srivastava et~al.(2022)Srivastava, Rastogi, Rao, Shoeb, Abid, Fisch, Brown, Santoro, Gupta, Garriga-Alonso et~al.}]{srivastava2022beyond}
Aarohi Srivastava, Abhinav Rastogi, Abhishek Rao, Abu Awal~Md Shoeb, Abubakar Abid, Adam Fisch, Adam~R Brown, Adam Santoro, Aditya Gupta, Adri{\`a} Garriga-Alonso, et~al. 2022.
\newblock Beyond the {I}mitation {G}ame: {Q}uantifying and {E}xtrapolating the {C}apabilities of {L}anguage {M}odels.
\newblock \emph{arXiv preprint arXiv:2206.04615}.

\bibitem[{Taori et~al.(2023)Taori, Gulrajani, Zhang, Dubois, Li, Guestrin, Liang, and Hashimoto}]{alpaca}
Rohan Taori, Ishaan Gulrajani, Tianyi Zhang, Yann Dubois, Xuechen Li, Carlos Guestrin, Percy Liang, and Tatsunori~B. Hashimoto. 2023.
\newblock Stanford alpaca: An instruction-following llama model.
\newblock \url{https://github.com/tatsu-lab/stanford_alpaca}.

\bibitem[{Touvron et~al.(2023)Touvron, Martin, Stone, Albert, Almahairi, Babaei, Bashlykov, Batra, Bhargava, Bhosale et~al.}]{touvron2023llama}
Hugo Touvron, Louis Martin, Kevin Stone, Peter Albert, Amjad Almahairi, Yasmine Babaei, Nikolay Bashlykov, Soumya Batra, Prajjwal Bhargava, Shruti Bhosale, et~al. 2023.
\newblock Llama 2: Open foundation and fine-tuned chat models.
\newblock \emph{arXiv preprint arXiv:2307.09288}.

\bibitem[{Wang et~al.(2024)Wang, Wei, Choi, and Ren}]{wang2024llms}
Siyuan Wang, Zhongyu Wei, Yejin Choi, and Xiang Ren. 2024.
\newblock \href {https://arxiv.org/abs/2402.11442} {Can llms reason with rules? logic scaffolding for stress-testing and improving llms}.
\newblock \emph{Preprint}, arXiv:2402.11442.

\bibitem[{Wang et~al.(2023)Wang, Ivison, Dasigi, Hessel, Khot, Chandu, Wadden, MacMillan, Smith, Beltagy et~al.}]{wang2023far}
Yizhong Wang, Hamish Ivison, Pradeep Dasigi, Jack Hessel, Tushar Khot, Khyathi~Raghavi Chandu, David Wadden, Kelsey MacMillan, Noah~A Smith, Iz~Beltagy, et~al. 2023.
\newblock How far can camels go? exploring the state of instruction tuning on open resources.
\newblock \emph{arXiv preprint arXiv:2306.04751}.

\bibitem[{Wang et~al.(2022{\natexlab{a}})Wang, Kordi, Mishra, Liu, Smith, Khashabi, and Hajishirzi}]{wang2022self}
Yizhong Wang, Yeganeh Kordi, Swaroop Mishra, Alisa Liu, Noah~A Smith, Daniel Khashabi, and Hannaneh Hajishirzi. 2022{\natexlab{a}}.
\newblock Self-instruct: Aligning language model with self generated instructions.
\newblock \emph{arXiv preprint arXiv:2212.10560}.

\bibitem[{Wang et~al.(2022{\natexlab{b}})Wang, Mishra, Alipoormolabashi, Kordi, Mirzaei, Naik, Ashok, Dhanasekaran, Arunkumar, Stap, Pathak, Karamanolakis, Lai, Purohit, Mondal, Anderson, Kuznia, Doshi, Pal, Patel, Moradshahi, Parmar, Purohit, Varshney, Kaza, Verma, Puri, Karia, Doshi, Sampat, Mishra, Reddy~A, Patro, Dixit, and Shen}]{wang-etal-2022-super}
Yizhong Wang, Swaroop Mishra, Pegah Alipoormolabashi, Yeganeh Kordi, Amirreza Mirzaei, Atharva Naik, Arjun Ashok, Arut~Selvan Dhanasekaran, Anjana Arunkumar, David Stap, Eshaan Pathak, Giannis Karamanolakis, Haizhi Lai, Ishan Purohit, Ishani Mondal, Jacob Anderson, Kirby Kuznia, Krima Doshi, Kuntal~Kumar Pal, Maitreya Patel, Mehrad Moradshahi, Mihir Parmar, Mirali Purohit, Neeraj Varshney, Phani~Rohitha Kaza, Pulkit Verma, Ravsehaj~Singh Puri, Rushang Karia, Savan Doshi, Shailaja~Keyur Sampat, Siddhartha Mishra, Sujan Reddy~A, Sumanta Patro, Tanay Dixit, and Xudong Shen. 2022{\natexlab{b}}.
\newblock \href {https://doi.org/10.18653/v1/2022.emnlp-main.340} {Super-{N}atural{I}nstructions: Generalization via declarative instructions on 1600+ {NLP} tasks}.
\newblock In \emph{Proceedings of the 2022 Conference on Empirical Methods in Natural Language Processing}, pages 5085--5109, Abu Dhabi, United Arab Emirates. Association for Computational Linguistics.

\bibitem[{Wettig et~al.(2024)Wettig, Gupta, Malik, and Chen}]{wettig2024qurating}
Alexander Wettig, Aatmik Gupta, Saumya Malik, and Danqi Chen. 2024.
\newblock \href {https://arxiv.org/abs/2402.09739} {Qurating: Selecting high-quality data for training language models}.
\newblock \emph{Preprint}, arXiv:2402.09739.

\bibitem[{Wu et~al.(2023)Wu, Waheed, Zhang, Abdul-Mageed, and Aji}]{wu2023lamini}
Minghao Wu, Abdul Waheed, Chiyu Zhang, Muhammad Abdul-Mageed, and Alham~Fikri Aji. 2023.
\newblock Lamini-lm: A diverse herd of distilled models from large-scale instructions.
\newblock \emph{arXiv preprint arXiv:2304.14402}.

\bibitem[{Xu et~al.(2023)Xu, Sun, Zheng, Geng, Zhao, Feng, Tao, and Jiang}]{xu2023wizardlm}
Can Xu, Qingfeng Sun, Kai Zheng, Xiubo Geng, Pu~Zhao, Jiazhan Feng, Chongyang Tao, and Daxin Jiang. 2023.
\newblock Wizardlm: Empowering large language models to follow complex instructions.
\newblock \emph{arXiv preprint arXiv:2304.12244}.

\bibitem[{Yehudai et~al.(2024)Yehudai, Carmeli, Mass, Arviv, Mills, Toledo, Shnarch, and Choshen}]{yehudai2024genie}
Asaf Yehudai, Boaz Carmeli, Yosi Mass, Ofir Arviv, Nathaniel Mills, Assaf Toledo, Eyal Shnarch, and Leshem Choshen. 2024.
\newblock \href {https://arxiv.org/abs/2401.14367} {Genie: Achieving human parity in content-grounded datasets generation}.
\newblock \emph{Preprint}, arXiv:2401.14367.

\bibitem[{Zellers et~al.(2019)Zellers, Holtzman, Bisk, Farhadi, and Choi}]{zellers2019hellaswag}
Rowan Zellers, Ari Holtzman, Yonatan Bisk, Ali Farhadi, and Yejin Choi. 2019.
\newblock \href {https://arxiv.org/abs/1905.07830} {Hellaswag: Can a machine really finish your sentence?}
\newblock \emph{Preprint}, arXiv:1905.07830.

\bibitem[{Zheng et~al.(2024{\natexlab{a}})Zheng, Mishra, Chen, Cheng, Chi, Le, and Zhou}]{zheng2024step}
Huaixiu~Steven Zheng, Swaroop Mishra, Xinyun Chen, Heng-Tze Cheng, Ed~H. Chi, Quoc~V Le, and Denny Zhou. 2024{\natexlab{a}}.
\newblock \href {https://arxiv.org/abs/2310.06117} {Take a step back: Evoking reasoning via abstraction in large language models}.
\newblock \emph{Preprint}, arXiv:2310.06117.

\bibitem[{Zheng et~al.(2024{\natexlab{b}})Zheng, Chiang, Sheng, Zhuang, Wu, Zhuang, Lin, Li, Li, Xing et~al.}]{zheng2024judging}
Lianmin Zheng, Wei-Lin Chiang, Ying Sheng, Siyuan Zhuang, Zhanghao Wu, Yonghao Zhuang, Zi~Lin, Zhuohan Li, Dacheng Li, Eric Xing, et~al. 2024{\natexlab{b}}.
\newblock Judging llm-as-a-judge with mt-bench and chatbot arena.
\newblock \emph{Advances in Neural Information Processing Systems}, 36.

\bibitem[{Zheng et~al.(2024{\natexlab{c}})Zheng, Guo, Qu, Guo, Zhang, Du, Lin, Huang, Chen, Fu et~al.}]{zheng2024kun}
Tianyu Zheng, Shuyue Guo, Xingwei Qu, Jiawei Guo, Weixu Zhang, Xinrun Du, Chenghua Lin, Wenhao Huang, Wenhu Chen, Jie Fu, et~al. 2024{\natexlab{c}}.
\newblock Kun: Answer polishment for chinese self-alignment with instruction back-translation.
\newblock \emph{arXiv preprint arXiv:2401.06477}.

\bibitem[{Zhou et~al.(2023)Zhou, Liu, Xu, Iyer, Sun, Mao, Ma, Efrat, Yu, Yu, Zhang, Ghosh, Lewis, Zettlemoyer, and Levy}]{zhou2023lima}
Chunting Zhou, Pengfei Liu, Puxin Xu, Srini Iyer, Jiao Sun, Yuning Mao, Xuezhe Ma, Avia Efrat, Ping Yu, Lili Yu, Susan Zhang, Gargi Ghosh, Mike Lewis, Luke Zettlemoyer, and Omer Levy. 2023.
\newblock \href {https://arxiv.org/abs/2305.11206} {Lima: Less is more for alignment}.
\newblock \emph{Preprint}, arXiv:2305.11206.

\bibitem[{Zhu et~al.(2024)Zhu, Zhang, Huang, Li, Niu, Fan, Lun, Tao, Su, Gong, Fang, and Liu}]{zhu2024plangpt}
He~Zhu, Wenjia Zhang, Nuoxian Huang, Boyang Li, Luyao Niu, Zipei Fan, Tianle Lun, Yicheng Tao, Junyou Su, Zhaoya Gong, Chenyu Fang, and Xing Liu. 2024.
\newblock \href {https://arxiv.org/abs/2402.19273} {Plangpt: Enhancing urban planning with tailored language model and efficient retrieval}.
\newblock \emph{Preprint}, arXiv:2402.19273.

\end{thebibliography}

\lstset{
  basicstyle=\ttfamily\small, 
  frame=single,
  frameround=fttt,
  breaklines=true,
  numbers=left,
  numberstyle=\tiny\color{gray}, 
  keywordstyle=\color{blue}, 
  showstringspaces=false
}

\clearpage
\newpage 

\appendix

\onecolumn

\section{Experiment Baselines} \label{appendix:baseline}

\begin{itemize}[leftmargin=0cm, itemindent=0.3cm]
    \item Alpaca-52k~\citep{alpaca}. This dataset is developed by Stanford University using Text-Davinci-003. It encompasses 52,002 instruction-following samples.
    \item Alpaca-GPT4~\citep{peng2023gpt4llm}. This dataset contains English Instruction-Following Data generated by GPT-4 using Alpaca prompts for fine-tuning LLMs. It encompasses 52,002 instruction-following samples, the same as Alpaca-52k.
    \item Alpaca-Cleaned. This is a cleaned version of the Alpaca-GPT4 Dataset to address problems like hallucinations, merged instruction, and so on. It encompasses 51,760 instruction-following samples.
    \item LIMA~\citep{zhou2023lima}.This is a dataset of 1,000 prompts and responses from a variety of sources, primarily split into community Q\&A forums and manually authored examples, where the outputs (responses) are stylistically aligned with each other, but the inputs (prompts) are diverse.
    \item WizardLM-70k~\citep{xu2023wizardlm}. This dataset employs the Evol-Instruct algorithm to enhance the quality of instruction data. Incorporating ChatGPT during the reformulation phase ensures  the data fidelity. Among its 250,000 instructions, we primarily focused on the WizardLM-7b subset, which consists of 70,000 samples.
    \item Muffin~\citep{lou2024muffin} MUFFIN's data curation includes input sampling, instruction collection via two methods, output annotation by ChatGPT/GP4-4, instruction filtering, and classification expansion. This is a large dataset of 68k training instances.
    \item ShareGPT~\citep{vicuna2023}. This is a human-annotated dataset consisting of approximately 70K user-shared conversations collected from ShareGPT.
    \item Humpback. This self-alignment method generates instruction data through reverse fine-tuning.
    
\end{itemize}

\section{\mname Details}

\subsection{Pre-screen Details} \label{appendix:pre-screen}
Our objective was to efficiently enhance the selection process, minimizing time spent while maximizing quality outcomes. Initially, we employed \textbf{Mistral-7b-instruct-v2} \citep{jiang2023mistral} to evaluate texts for repetitive content, personal privacy concerns, specific themes, and advertising, using prompts to guide scoring and annotation (see Table \ref{table:text_quality_prompt_1}). For diversity assessment, we utilized a fast community detection algorithm \ref{algorithm:fast_community_detection} with hyperparameters set to k = 2 and sim\-ratio = 0.7(k: the minimum size of a community; sim\-ratio: controls the similarity threshold, Only node pairs with similarity scores higher than this threshold are considered connected), facilitating the classification of half a million entries within minutes. The model paraphrase-MiniLM-L6-v2 \citep{reimers-2019-sentence-bert}) is used for text embedding. For larger datasets, texts were segmented into groups for individual community detection analyses. After the pre-screening process, \textit{Pre-Screen Data} has approximately 30k records, which is 6\% of the original.
This stage was designed to balance the trade-off between processing speed and analytical precision, prioritizing efficiency over exhaustive detail examination.

\subsection{UCB Bootstrap} \label{appendix:UCB}
The setup comprises a language model \(G\) parameterized by \(\theta_G\) for generating instructions, a critic model \(J\) parameterized by \(\theta_J\) for evaluating instruction quality, as well as a document set \(\mathcal{D}\), a subset \(\mathcal{D^{'}}\), task-type tags \(\mathcal{T_T}\), and difficulty-level tags \(\mathcal{T_D}\).

The procedure is as follows:

\begin{enumerate}[left=0em]
    \item \textbf{Initialization:} 
    \begin{align*}
        S &\leftarrow \emptyset \\
    \end{align*}
    
    \item \textbf{Seed Generation (\(SeedGen\)):}
    \begin{align*}
        \forall d \in \mathcal{D^{'}}, \ &generate\ x_i \sim P(x | d, t; \theta_G) \\
        &where\ t \sim \mathcal{U}(\mathcal{T_T} \times \mathcal{T_D}) \\
        &S \leftarrow S \cup \{x_i\}
    \end{align*}
    
    \item \textbf{Instruction Augmentation (\(InsAug\)):}
    For \(f\) rounds or until \(|S|\) reaches a desired threshold:
    \begin{enumerate}[label=\alph*.,left=0em]
        \item Select a subset \(S' \subset S\) using the UCB strategy:
        \begin{align*}
            UCB(s) &= \bar{x}_s + C\sqrt{\frac{2 \ln N}{n_s}} \\
            S' &= \{s_i | s_i \in S, UCB(s_i) \text{ is maximized}\}
        \end{align*}
        where \(\bar{x}_s\) is the average quality score of instruction \(s\), \(N\) is the total number of selections, \(C\) is a hyper-parameter constant used to control exploration, t and \(n_s\) is the number of times instruction \(s\) has been selected.
        
        \item For each \(s_i \in S'\), generate augmented instructions \(x'\):
        \begin{align*}
            x' &\sim P(x | c, S'; \theta_G) \\
            &\text{s.t. } Sim(x'; s_i) < \tau
        \end{align*}
        where \(\tau\) is a similarity threshold.
        
        \item Update \(S\) with the augmented instructions:
        \begin{align*}
            S &\leftarrow S \cup \{x'\}
        \end{align*}
    \end{enumerate}
\end{enumerate}

\subsection{Fast Community Detection Algorithm} \label{appendix:fast_community_detection_algorithm}
As Algorithm \ref{algorithm:fast_community_detection} has shown, the Fast Community Detection Algorithm is used to cluster the embeddings of instructions processed by SentenceTransformer \citep{reimers-2019-sentence-bert}, which can then represent the diversity of instructions. 
Specifically, Fast Community Detection works by iteratively identifying groups of data points (embeddings of sentences) that are closely related based on a predefined similarity threshold, efficiently leveraging cosine similarity calculations. It prioritizes larger communities while minimizing overlapping clusters to produce meaningful community structures.



\begin{algorithm*}
\caption{Fast Community Detection \citep{reimers-2019-sentence-bert}}
\label{algorithm:fast_community_detection}
\begin{algorithmic}[1]
\Function{CommunityDetection}{$\text{embeddings}, \text{threshold}, \text{min\_community\_size}, \text{batch\_size}$}
    \State Normalize $\text{embeddings}$
    \State Initialize $\text{extracted\_communities}$ as empty list

    \For{$\text{start\_idx}$ \textbf{in} range(0, length(embeddings), batch\_size)}
        \State Compute cosine similarity scores for batch starting from $\text{start\_idx}$
        \State Find top-k values from cosine similarity scores
        \For{$i$ \textbf{in} range(length(top\_k\_values))}
            \If{last element of $i$-th top-k values $\geq$ $\text{threshold}$}
                \State Find top-k most similar entries for $i$-th element
                \While{last element of top-k values $>$ $\text{threshold}$ \textbf{and} $\text{sort\_max\_size} <$ length of $\text{embeddings}$}
                    \State Increase $\text{sort\_max\_size}$ if needed
                \EndWhile
                \State Add indices of entries with similarity $\geq$ $\text{threshold}$ to $\text{extracted\_communities}$
            \EndIf
        \EndFor
    \EndFor
    
    \State Sort $\text{extracted\_communities}$ by size
    \State Remove overlapping communities from $\text{extracted\_communities}$
    
    \State \textbf{return} $\text{extracted\_communities}$
\EndFunction
\end{algorithmic}
\end{algorithm*}

\section{Experiment Setting Detail}
\label{appendix:setting}
We chose LoRA over full fine-tuning due to similar performance observed in preliminary experiments, with computational constraints being the primary factor influencing this decision. 

We use the same hyperparameters as existing supervised instruction tuning methods~\citep{vicuna2023,raschka_lora_2023}. Specifically, we use cosine learning rate scheduling with a starting learning rate of $2 \times 10^{-5}$ and a weight decay of 0.1. The batch size is 32 and the dropout rate is 0.1. For the LoRA configuration, we employ a rank of 256 and set $\alpha$ to 512, with an initial learning rate of $5 \times 10^{-5}$. We utilize 8 NVIDIA 4090 GPUs to train our model.


\section{Prompt Templates Used in \mname}
\subsection{Text Filtering}
\label{appendix:text_filtering}

\newenvironment{lmttfont}{\ttfamily}{}
\begin{table*}[h]
\caption{
Prompts for Pre-Screen
\label{table:text_quality_prompt_1}
}
\setstretch{0.7}
\fbox{
\begin{minipage}{40em}
\begin{small}
\begin{lmttfont}
You are act as a assistant to check useless, informal or ambiguous information. Let's think step by step.\\
The objective is to meticulously inspect the text to determine if it is useless, informal or ambiguous text (e.g. random characters, ambiguous paragraph, broken sequence, informally organized text, etc.)\\
Your response should be '1' (yes) if the text contains useless, informal or ambiguous information, or '0' (no) if it does not, without providing any reasoning and explanation.\\
\\
\#\#\# Document:\\
\{doc\}\\
\\
\#\#\# Answer:\\
\end{lmttfont}
\end{small}
\end{minipage}
}

\fbox{
\begin{minipage}{40em}
\begin{small}
\begin{lmttfont}
You are act as a assistant to check privacy information. Let's think step by step.\\
The objective is to meticulously inspect the text to determine if it contains any privacy information (e.g. human names, phone numbers, addresses, etc.).\\
Your response should be '1' (yes) if the text contains privacy information, or '0' (no) if it does not, without providing any reasoning and explanation.\\
\\
\#\#\# Text:\\
\{doc\}\\
\\
\#\#\# Answer:\\
\end{lmttfont}
\end{small}
\end{minipage}
}

\fbox{
\begin{minipage}{40em}
\begin{small}
\begin{lmttfont}
I want you to act as an advertisement evaluator. Let's think step by step.\\
The objective is to meticulously inspect the text based on certain characteristics and decide whether it is an advertisement or not.\\
Your response should be '1' (yes) if the text is an advertisement, or '0' (no) if it is not, without providing any reasoning and explanation.\\
\\
Evaluate the text considering these characteristics:\\
\\
- Promotional language or sales pitch\\
- Mention of product or service benefits\\
- Call to action (e.g., "Buy now", "Subscribe")\\
- Pricing information or special offers\\
- Contact information or links for more details\\
\\
\textless Answer Format\textgreater: 1 or 0\\
\\
\#\#\# Text:\\
\{text\}\\
\\
\#\#\# Answer:\\
\end{lmttfont}
\end{small}
\end{minipage}
}
\vspace{1mm}
\end{table*}

\begin{table*}[h]

\caption{
Prompts for instruction generation filter
\label{table:text_quality_prompt_2}
}
\setstretch{0.7}

\fbox{
\begin{minipage}{40em}
\begin{small}
\begin{lmttfont}
I want you to act as an instruction evaluator. Please evaluate this instruction and respond with '0' (bad) or '1' (good), without giving reasons.\\
Standard: A good instruction Must not involve recent or current events. Historical events are fine.\\
Example1:\\
Instruction: Please analyze the recent COVID-19 outbreak.\\
Answer: 0 (Reason: recent)\\
Example2:\\
Instruction: What's happening in China in September 2023?\\
Answer: 0 (Reason: in September 2023)\\
Example3:\\
Instruction: Provide an account of events from last Monday night.\\
Answer: 0 (Reason: last Monday night)\\
\\
\#\#\# Instruction:\\
\{instruction\}\\
\#\#\# Answer: \\
\end{lmttfont}
\end{small}
\end{minipage}
}
\fbox{
\begin{minipage}{40em}
\begin{small}
\begin{lmttfont}
I want you to act as a instruction evaluator. Please evaluate this instruction and respond with '0' (bad) or '1' (good), without giving reasons.\\
Standard: A good instruction must not include any private information like names, addresses, phone numbers, etc, unless the person is historical or famous.\\
Example1:\\
Instruction: What is the name of the person who lives at 123 Main Street?\\
Answer: 0 (Reason: private information)\\
Example2:\\
Instruction: What is the name of the first president of the United States?\\
Answer: 1 (Reason: historical)\\
Example3:\\
Instruction: What is the address of the CEO of Microsoft?\\
Answer: 0 (Reason: private information)\\
\\
\#\#\# Instruction:\\
\{instruction\}\\
\#\#\# Answer: \\
\end{lmttfont}
\end{small}
\end{minipage}
}

\fbox{
\begin{minipage}{40em}
\begin{small}
\begin{lmttfont}
I want you to act as a instruction evaluator. Please evaluate this instruction and respond with '0' (bad) or '1' (good), without giving reasons.\\
Standard: A good instruction is perfectly logical, and practical, and can be fully understood by a human.\\
A bad instruction, likely generated by AI, is generally vague, weird, complex, and long. It may seem to string unrelated words, topics, and tasks together.\\
Example1: \\
Instruction: Considering the health benefits of a non-dairy diet, how does the emotional response of individuals vary when they attend social events where dairy-based foods are served?\\
Answer: 0\\
Example2:\\
Instruction: Create a multidisciplinary essay that explores the and historical origins of the dish 'Shrimp Alfredo Pasta Bake'. Discuss the various ingredients, their origins. Additionally, translate the recipe instructions from English to Spanish.\\
Answer: 0\\
\\
\#\#\# Instruction:\\
\{instruction\}\\
\#\#\# Answer:\\
\end{lmttfont}
\end{small}
\end{minipage}
}

\vspace{1mm}
\end{table*}


\clearpage
\newpage

Table \ref{table:text_quality_prompt_1} shows the prompts for basic filtering, including filtering information with useless information, privacy information, or advertisement. 

Table \ref{table:text_quality_prompt_2} shows the prompts for instruction generation filtering, including filtering instructions that are time-sensitive, asking for private information, or not answerable.

\subsection{Complexity and Quality Scorer}
\begin{table}[htbp]
\caption{
Prompt for quality scorer
\label{table:text_quality_scorer_prompt}
}
\fbox{
\begin{minipage}{40em}
\begin{small}
\begin{lmttfont}
You are a helpful assistant. Please identify the quality score of the Response corresponding to the Question.\\
\#\#\# Question:\\
\{instruction\}\\
\#\#\# Response:\\
\{output\}\\
\#\#\# Quality:\\
\end{lmttfont}
\end{small}
\end{minipage}
}
\vspace{1mm}
\end{table}

\begin{table}[thbp]
\caption{
Prompt for complexity scorer
\label{table:text_complexity_scorer_prompt}
}
\fbox{
\begin{minipage}{40em}
\begin{small}
\begin{lmttfont}
You are a helpful assistant. Please identify the complexity score of the following user query.\\
\#\#\# Query:\\
\{instruction\}\\
\#\#\# Complexity:\\
\end{lmttfont}
\end{small}
\end{minipage}
}
\vspace{1mm}
\end{table}

As Table \ref{table:text_quality_scorer_prompt} and \ref{table:text_complexity_scorer_prompt} have shown, the prompts are provided to deita-complexity-scorer and deita-quality-scorer model \citep{liu2023what}.

\subsection{Generating Instruction Pairs}
\label{appendix:response_generate}
\mname employs 2 ways to generate instruction response:
\begin{itemize}
    \item Question, Document to Answer: model infers answer with both question and related document.
    \item Question to Answer: model infers the answer directly with the question, using its own knowledge.
\end{itemize}
\begin{table}[htbp]
\caption{Question, Document to Answer}
\label{table:qdoc2a}
\fbox{
\begin{minipage}{40em}
\begin{small}
\begin{lmttfont}
You are a helpful, respectful and honest assistant. Always answer as helpfully as possible, while being safe. Your answers should not include any harmful, unethical, racist, sexist, toxic, dangerous, or illegal content. Please ensure that your responses are socially unbiased and positive in nature.\\
If a question does not make any sense, or is not factually coherent, explain why instead of answering something not correct. If you don't know the answer to a question, please don't share false information.\\
\#\#\# Instruction: \{question\}.\\
\#\#\# Paragraph: \{doc\}.\\
\#\#\# Response:
\end{lmttfont}
\end{small}
\end{minipage}
}
\vspace{1mm}
\end{table}



\begin{table}[htbp]
\caption{Question to Answer}
\label{table:q2a}
\fbox{
\begin{minipage}{40em}
\begin{small}
\begin{lmttfont}
You are a helpful, respectful and honest assistant. Always answer as helpfully as possible, while being safe. Your answers should not include any harmful, unethical, racist, sexist, toxic, dangerous, or illegal content. Please ensure that your responses are socially unbiased and positive in nature.\\
If a question does not make any sense, or is not factually coherent, explain why instead of answering something not correct. If you don't know the answer to a question, please don't share false information.\\
\#\#\# QUESTION: \{question\}\\
\#\#\# Response:
\end{lmttfont}
\end{small}
\end{minipage}
}
\vspace{1mm}
\end{table}

\subsection{Seed Generation} \label{sec:app-seed-gen}
\begin{lstlisting}[language=Python, label={lst:generate_prompt_with_tags}, caption={Seed Generation}]
def seed_gen(text):
    reasoning_tag = "It should be complex and requires multiple-step reasoning to solve."
    critical_thinking_tag = "It demands critical thinking skills to analyze from various perspectives and evaluate multiple solutions."
    creativity_tag = "It necessitates creative thinking to devise innovative solutions beyond conventional approaches."
    interdisciplinary_tag = "It demands integrating knowledge from diverse disciplines to address its multifaceted nature."
    command_tag = "It should be in the style of a command or imperative. For example, 'Write a paragraph about...' or 'Describe the...'"
    question_tag = "It should be in the style of a question or interrogative. For example, 'What is the...?' or 'How do you...?'"
    
    nli_tag = "It is a Natural language inference question: Assessing if evidence supports a conclusion."
    commonsense_tag = "It is a Commonsense question: Predicting outcomes based on everyday knowledge."
    sentiment_tag = "It is a Sentiment analysis question: Determining emotional response to a given scenario."
    paraphrase_tag = "It is a Paraphrasing question: Rewording a statement while retaining its meaning."
    close_book_qa_tag = "It is a Close-book QA question: Answering factual queries using pre-existing knowledge."
    struc2text_tag = "It is a Structure to text question: Describing a process or concept in written form."
    summarization_tag = "It is a Summarization question: Condensing key information from a larger text."
    translate_tag = "It is a Translation question: Converting text from one language to another."
    implicit_reasoning_tag = "It is a Implicit reasoning question: Inferring reasons behind common behaviors."
    text_category_tag = "It is a Text categorization question: Identifying defining characteristics of a given text type."
    
    tags = [reasoning_tag, critical_thinking_tag, creativity_tag, interdisciplinary_tag]
    classify = [nli_tag, commonsense_tag, sentiment_tag, paraphrase_tag, close_book_qa_tag, struc2text_tag, summarization_tag, translate_tag, implicit_reasoning_tag, text_category_tag]
    types = [command_tag, question_tag]
    
    QUESTION_TEMPLATE = """You're proficient in crafting complex question. Generate only one question that adheres to the provided #Paragraph#.
    The question should meet the following criteria:
    0. The person answering the question cannot see the #Paragraph#[SYSTEM: IMPORTANT], so the question must not contain phrases like 'Given the information provided', 'Based on the provided information', or similar expressions that imply direct citations or references from #Paragraph#.
    1. {characteristic}.
    2. {type}.
    3. {classify}.

    ### Paragraph:
    {text}
    ### Question:
    """
    prompts = [QUESTION_TEMPLATE.format(characteristic=tag, type=type, text=text, classify=c) for tag in tags for c in classify for type in types]
    return prompts
\end{lstlisting}
Code \ref{lst:generate_prompt_with_tags} shows the process of generating seed with sampled tags, including task types and difficulty levels.

\subsection{Think Different Prompt}\label{sec:app-think-differently}
\begin{table}[htbp]
\caption{Prompt for Think Differently}
\label{table:ucb_think_diff}
\fbox{
\begin{minipage}{40em}
\begin{small}
\begin{lmttfont}
You are a helpful assistant. Your task is to conceive a complex question inspired from the Paragraph, while ensuring it is completely different from the example provided below. Prohibit the use of expressions, question types, and initial verbs that are identical to those in the Examples provided. Avoid phrases such as 'Based on', 'Given the information provided', 'Using the data' or any similar expressions that suggest references to the Paragraph.
{command}

\#\#\# Counterexample: \\
<Example1>: \{seed1\} \\
<Example2>: \{seed2\} \\
<Example3>: \{seed3\} \\
<Example4>: \{seed4\} \\
<Example5>: \{seed5\} \\

\#\#\# Paragraph: \\
\{text\} \\

\#\#\# Question:
\end{lmttfont}
\end{small}
\end{minipage}
}
\vspace{1mm}
\end{table}

\subsection{Self-Instruct Prompting Templates for Data Generation}
\label{sec:app-self-instruct-gen}

\textit{Self-Instruct} relies on the following prompting template in order to elicit the generation from language models.



\newcommand{\pa}[1]{ {\color{blue}\{} #1 {\color{blue}\}} }

\begin{table*}[ht]
    \centering
    \small
    \noindent\fbox{%
    \begin{minipage}{\dimexpr\textwidth-2\fboxsep-2\fboxrule} 
\tt 
Come up with a series of tasks:\\
\\
Task 1: \{instruction for existing task 1\} \\
Task 2: \{instruction for existing task 2\} \\
Task 3: \{instruction for existing task 3\} \\
Task 4: \{instruction for existing task 4\} \\
Task 5: \{instruction for existing task 5\} \\
Task 6: \{instruction for existing task 6\} \\
Task 7: \{instruction for existing task 7\} \\
Task 8: \{instruction for existing task 8\} \\
Task 9:
    \end{minipage}
}

    \caption{Prompt used for \textit{Self-Instruct}}
    \label{tab:instruction_generation_template}
\end{table*}

\subsection{Faithfulness Evaluation}
\begin{table}[htbp]
\caption{
Prompt for Faithfulness Evaluation \citep{li2024selfalignment}
\label{table:faithfulness_evaluation_prompt}
}
\fbox{
\begin{minipage}{40em}
\begin{small}
\begin{lmttfont}
Below is an instruction from an user and a candidate answer. 
Let's think step by step.
Evaluate whether or not the answer is a good example of how AI Assistant should respond to the user's instruction. Please assign a score using the following 5-point scale:
1: It means the answer is incomplete, vague, off-topic, or not exactly what the user asked for. For example, some content seems missing. Or the response is from another person’s perspective with their personal experience (e.g. taken from blog posts). Or it contains promotional text or other irrelevant information.
2: (between 1 and 3)
3: It means the answer is helpful but not written by an AI Assistant. It addresses all the basic asks from the user. It is complete and self contained with the drawback that the response is not written from an AI assistant's perspective, but from other people's perspective. For example, it contains personal experience or opinion, mentions comments section, or share on social media, etc.
4: (between 3 and 5)
5: It means it is a perfect answer from an AI Assistant. It has a clear focus on being a helpful AI Assistant, where the response looks like intentionally written to address the user's question or instruction without any irrelevant sentences. The answer provides high quality content, demonstrating expert knowledge in the area, is very well written, logical, easy-to-follow, engaging and insightful.

Your reply should be only 1 or 2 or 3 or 4 or 5, without providing any reasoning and explanation.

\#\#\# Instruction:
\{instruction\}

\#\#\# Answer:
\{response\}

\#\#\# Your Reply:
\end{lmttfont}
\end{small}
\end{minipage}
}
\vspace{1mm}
\end{table}

Table \ref{table:faithfulness_evaluation_prompt} shows the prompt to select more faithful instruction. The prompt originates from \citep{li2024selfalignment} with minor modifications.

\newpage
\section{Data Analysis}
\subsection{Quality, Length and Diversity}
\label{appendix:quality_length_diversity}

\begin{figure}[h]
    \begin{minipage}[b]{0.45\textwidth}
        \centering
        \includegraphics[width=1\textwidth]{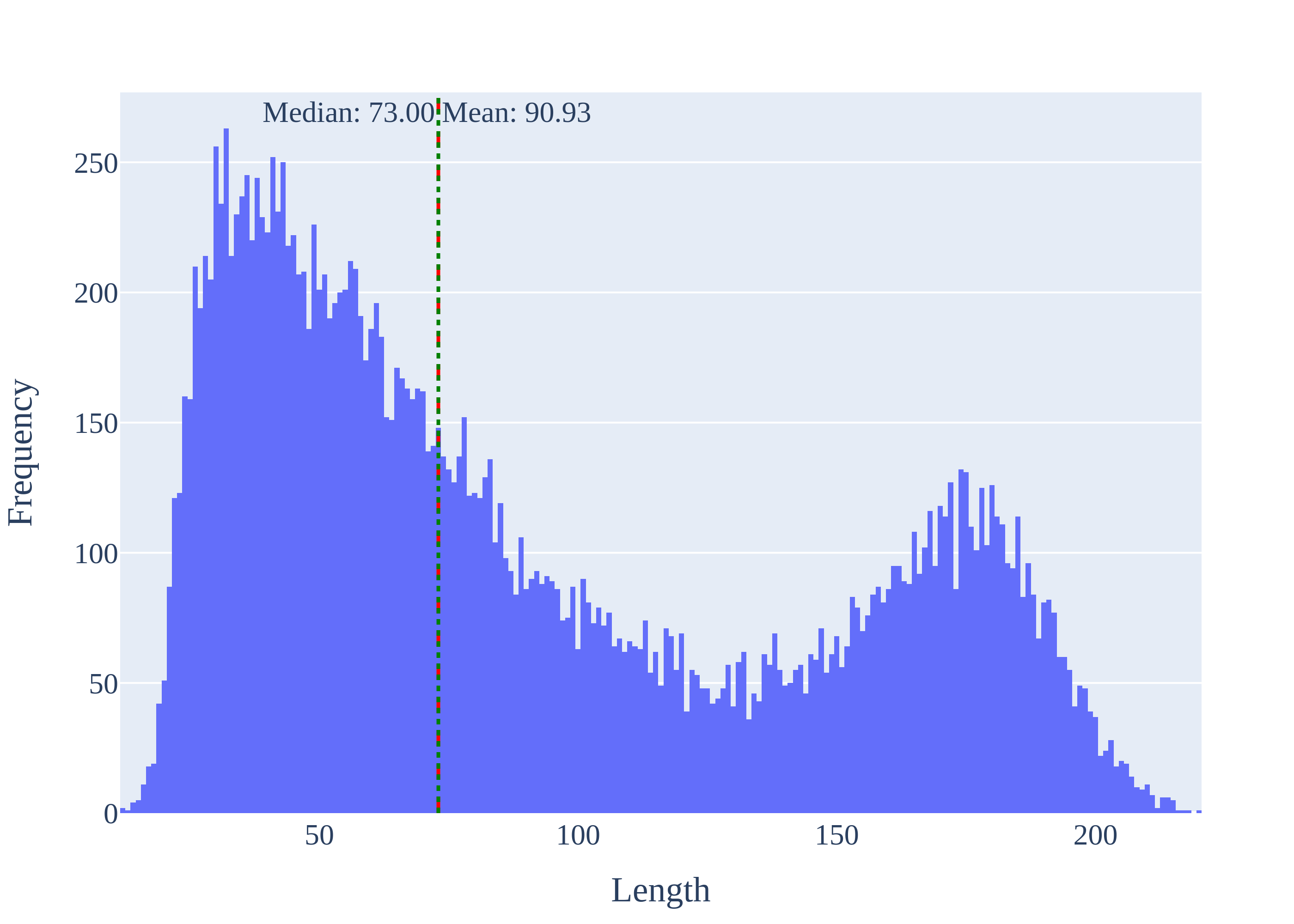}
        \caption{\mname Instruction Length Distribution}
        \label{fig:fanno_instruction_length_distribution}
    \end{minipage}
    \hfill
    \begin{minipage}[b]{0.45\textwidth}
        \centering
        \includegraphics[width=1\textwidth]{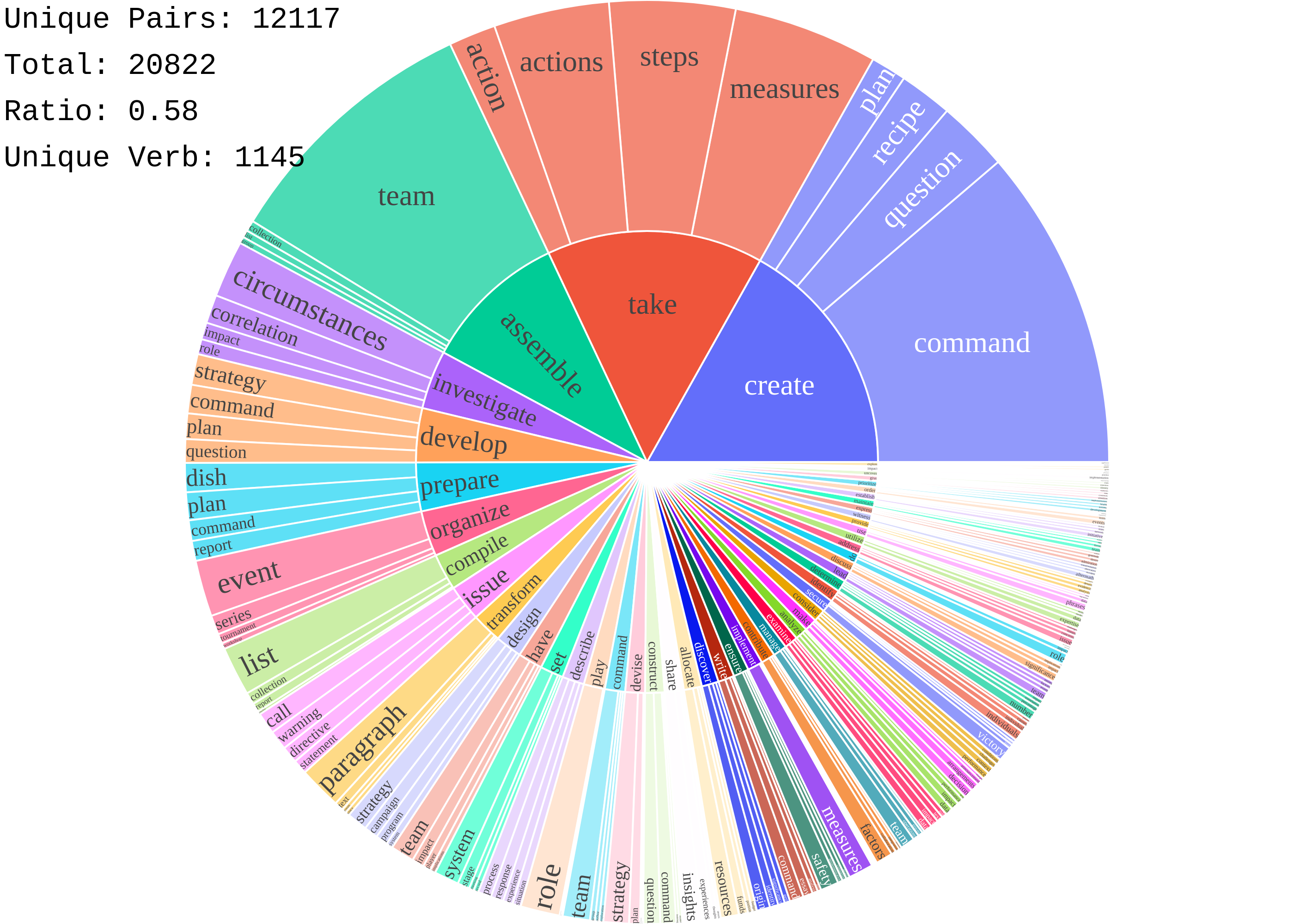}
        \caption{Top 50 common verbs and their corresponding nouns in \mname}
        \label{fig:fanno_instruction_verb_noun}
    \end{minipage}
\end{figure}

\begin{figure}[h]
    \begin{minipage}[b]{0.45\textwidth}
        \centering
        \includegraphics[width=1\textwidth]{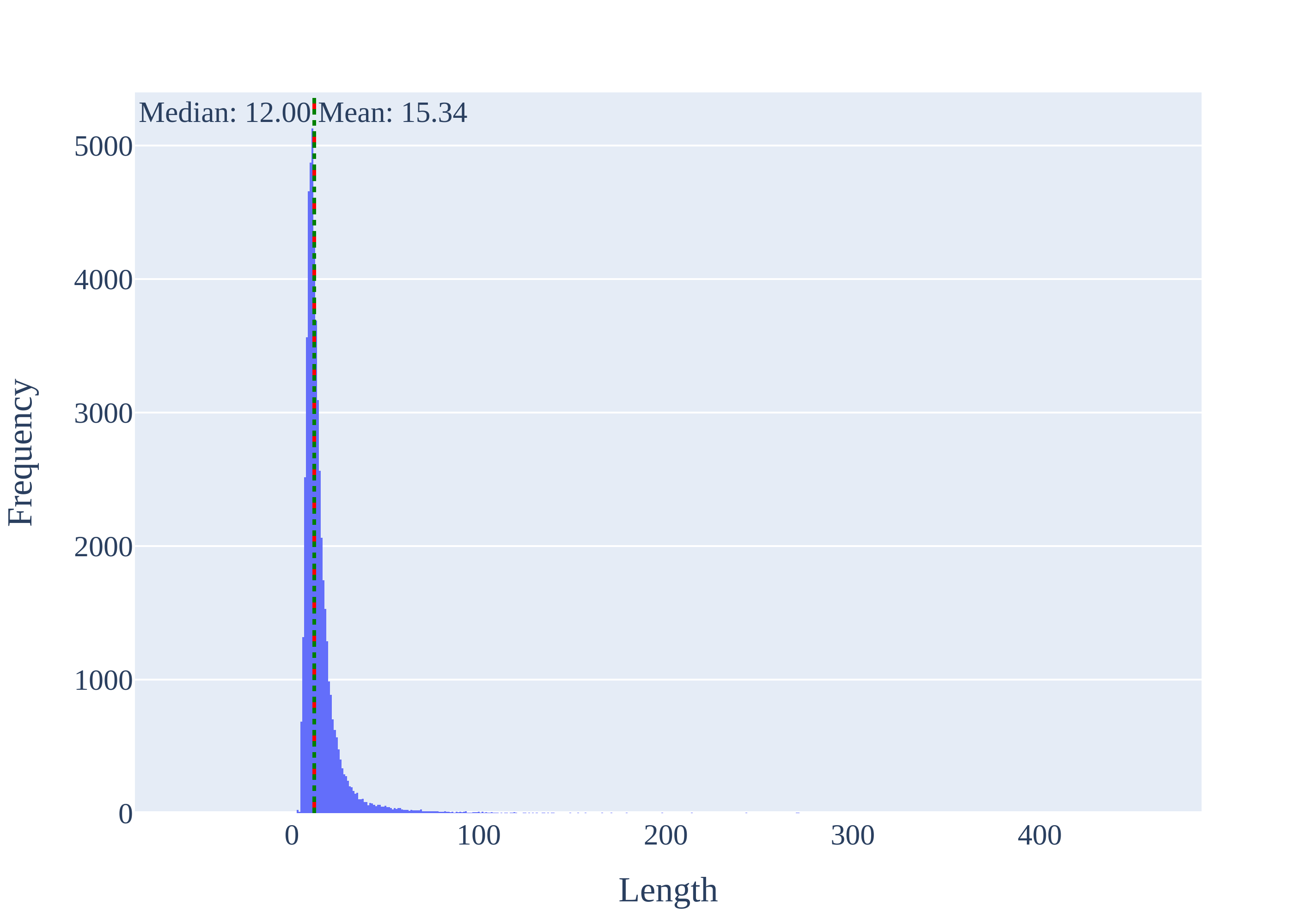}
        \caption{Alpaca-Cleaned Instruction Length Distribution}
        \label{fig:alpaca_instruction_length_distribution}
    \end{minipage}
    \hfill
    \begin{minipage}[b]{0.45\textwidth}
        \centering
        \includegraphics[width=1\textwidth]{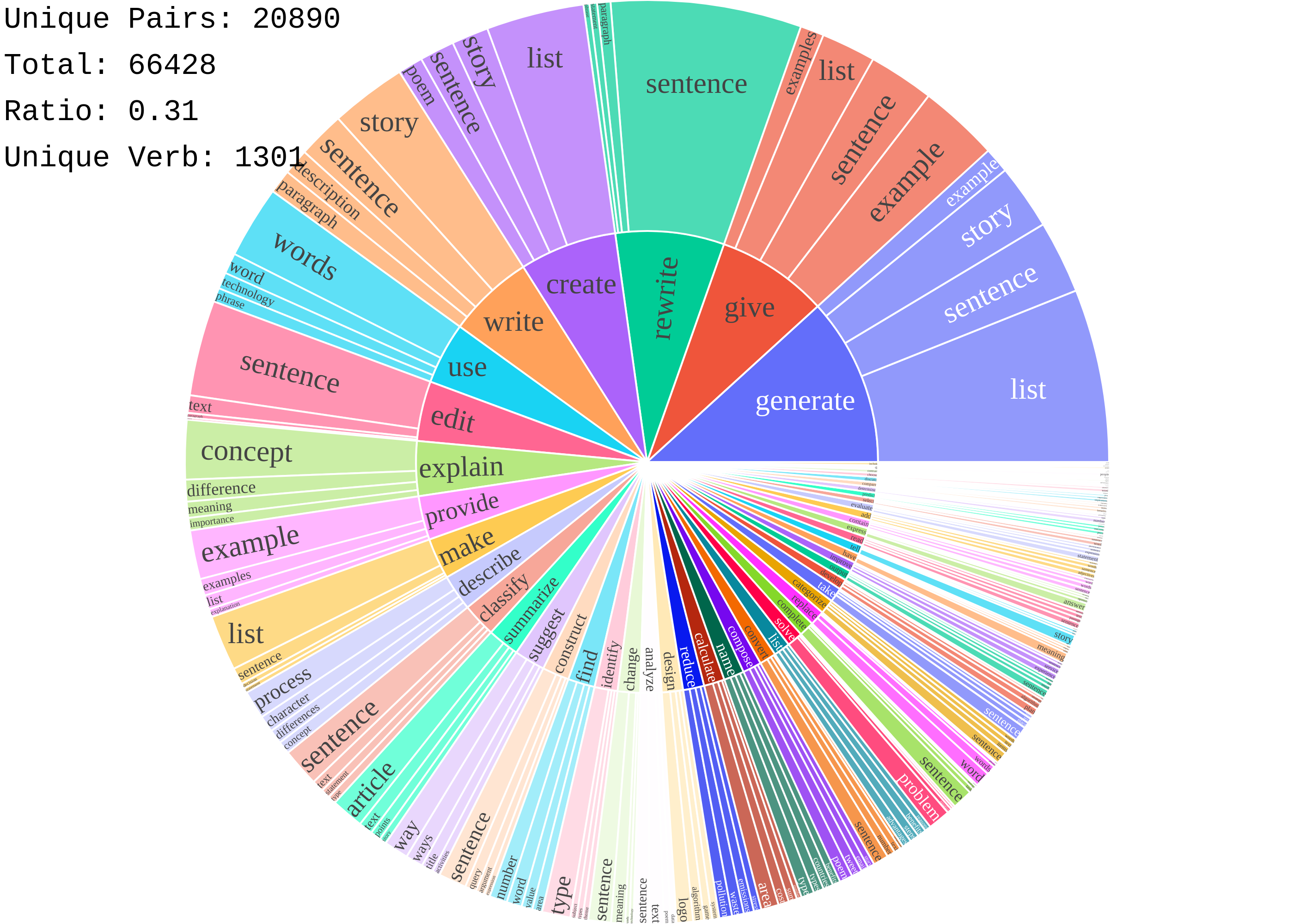}
        \caption{Top 50 common verbs and their corresponding nouns in Alpaca-Cleaned}
        \label{fig:alpaca_clean_instruction_verb_noun}
    \end{minipage}
\end{figure}

\begin{figure}
    \centering
    \includegraphics[width=0.5\textwidth]{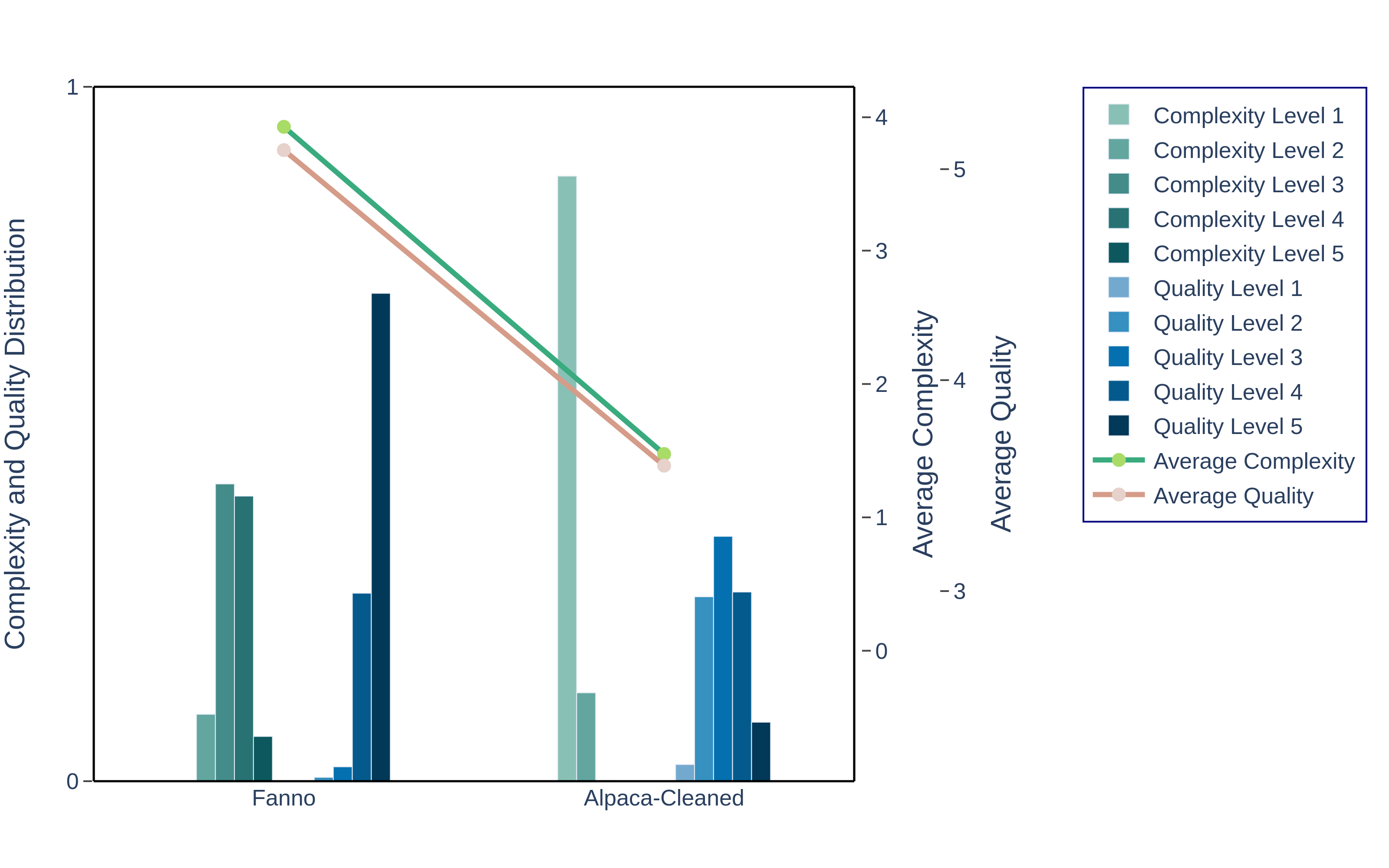}
    \caption{Quality and Complexity Comparison between \mname and Alpaca-Cleaned}
    \label{fig:quality_complexity}
\end{figure}

Figures~\ref{fig:fanno_instruction_length_distribution} and \ref{fig:alpaca_instruction_length_distribution} show the instruction length distribution of \mname and Alpaca-Cleaned, respectively. It is worth noting that the mentioned length includes both the instruction and input combined. Figures~\ref{fig:fanno_instruction_verb_noun} and \ref{fig:alpaca_clean_instruction_verb_noun} show the verb-noun diversity of \mname and Alpaca-Cleaned, respectively. Figure~\ref{fig:quality_complexity} shows the comparison of quality and complexity between \mname and Alpaca-Cleaned.

\newpage
\section{Human Evaluation}

\subsection{Complexity Level}
The first tier (0 point) pertains to instructions that exhibit apparent issues, such as being unanswerable or containing missing information. The second tier (1 point) involves instructions that can be answered using everyday knowledge. These instructions may assess basic skills, analyze human emotional experiences, or organize activities without requiring much specialized knowledge in a particular field. The third tier (2 points) comprises expert instructions. These necessitate specialized knowledge and require thorough deliberation steps to fulfill the instruction's requirement. 

\subsection{Instruction Complexity Human Evaluation} \label{appendix:human_eval_complexity}
\subsubsection*{2 point (Expert Level)}
\begin{enumerate}[label=\arabic*.]
    \item Create a series of interactive exercises for a group of advanced French learners to practice the conditional tense, incorporating a variety of verb forms and sentence structures, while also encouraging them to engage in peer-to-peer learning and problem-solving. Consider using a combination of written and oral activities, and provide clear instructions and examples for each exercise. Additionally, design a system for assessing their progress and providing personalized feedback.
    \item How can we optimize the WordPress website's performance for logged-in users without employing the Auto-Cache Engine? Consider various caching strategies and evaluate their potential impact on user experience and website functionality.
    \item Design a multifaceted approach to streamline the patient registration process for a healthcare facility, ensuring adherence to ICD-10 and CPT coding standards, while providing exceptional customer service to a diverse patient population. Consider implementing innovative technologies and collaborating with various departments to optimize workflows and enhance overall efficiency. Evaluate the potential impact of this approach on patient satisfaction, staff morale, and financial performance.
    \item Assemble a team of data experts to evaluate the potential impact of a centralized data strategy on the decision-making process of a tech startup, considering the long-term benefits and potential drawbacks. Analyze various case studies of successful companies, such as Google, Apple, Amazon, and Facebook, to identify key strategies and best practices for implementing a data-driven culture. Evaluate the role of immediate returns versus long-term benefits in the adoption of data-driven decision-making and provide recommendations for managing potential challenges, such as data security and privacy concerns.
    \item Assemble a team of nutritionists and chefs to devise a creative and nutritious menu for a charity gala, utilizing natural sweeteners as the primary ingredient in each dish, while ensuring that the final creations are visually appealing and can be prepared in large quantities. Additionally, consider the dietary restrictions of various attendees and incorporate alternative options for those with gluten, dairy, and nut allergies. The team should also aim to minimize food waste and maximize the use of locally sourced ingredients.
\end{enumerate}

\subsubsection*{1 point (Everyday Level)}
\begin{enumerate}[label=\arabic*.]
    \item Develop a weekly routine that integrates both your professional and personal commitments, ensuring that you effectively manage your time and accomplish your goals. What unique strategies could you employ to optimize your productivity during your weekly review and planning session?
    \item Persuade your employer to grant you the flexibility to work from home for a specified number of days per week, demonstrating the potential time and cost savings, as well as the potential benefits to your overall well-being.
    \item Capture the essence of a cherished memory by taking a photograph of a cherished photograph. Ensure the image is visually appealing and evokes a sense of nostalgia.
    \item Translate the following paragraph from English to another language of your choice. Ensure that the translation conveys the original meaning and intent.
    "Analyze the artworks displayed at the exhibition from various perspectives. Which artwork resonates the most with the theme of environmental conservation? Provide reasons for your answer."
\end{enumerate}

\subsubsection*{0 point (Bad)}
\begin{enumerate}[label=\arabic*.]
    \item Utilize the data from the Neighbourhood Forum Launch event to determine the percentage of attendees who were members prior to the event and the percentage who joined during the event. Additionally, identify the top three focus groups with the highest number of attendees and determine the average number of attendees per focus group. Finally, calculate the total number of attendees who placed a dot on the Forum map and the percentage of attendees who did so.
\end{enumerate}

\end{document}